\documentclass[letterpaper, 10 pt, conference]{ieeeconf}  

\IEEEoverridecommandlockouts                              

\usepackage[pdftex]{graphicx}
\usepackage{amssymb}
\usepackage{amsmath}
\usepackage{bm}
\usepackage{subfigure}
\usepackage{cite}
\usepackage{epsfig}
\usepackage{caption}
\usepackage{color}
\usepackage[dvipsnames]{xcolor}
\usepackage{psfrag}
\usepackage{cases}
\usepackage{acronym}
\usepackage{setspace}
\usepackage{times}
\usepackage{latexsym}
\usepackage{dblfloatfix}
\usepackage[ruled,linesnumbered]{algorithm2e}
\usepackage{metalogo}
\usepackage{tabularx}
\usepackage{array}
\usepackage{multirow}
\usepackage{booktabs}
\usepackage{graphicx}

\let\labelindent\relax
\usepackage{enumitem}

\usepackage{hyperref}
 \usepackage{threeparttable}
\hypersetup{
    colorlinks=true,
    linkcolor=blue,
    filecolor=blue,      
    urlcolor=blue,
    citecolor=cyan,
}
\usepackage{geometry}
\graphicspath{{pics/}}

\title{\LARGE \bf
A Coarse-to-Fine Place Recognition Approach using Attention-guided Descriptors and Overlap Estimation\vspace{-3mm}
}

\author{Chencan Fu, Lin Li, Jianbiao Mei, Yukai Ma, Linpeng Peng, Xiangrui Zhao, and Yong Liu$^{*}$\vspace{-3mm}
\thanks{Chencan Fu, Lin Li, Jianbiao Mei, Yukai Ma, Linpeng Peng, Xiangrui Zhao, and Yong Liu are with the Institute of Cyber-Systems and Control, Zhejiang University, Hangzhou 310027, P. R. China. (*Yong Liu is the corresponding author, email: yongliu@iipc.zju.edu.cn).}
}

\begin{document}

\maketitle
\thispagestyle{empty}
\pagestyle{empty}

\begin{abstract}
Place recognition is a challenging but crucial task in robotics. Current description-based methods may be limited by representation capabilities, while pairwise similarity-based methods require exhaustive searches, which is time-consuming. In this paper, we present a novel coarse-to-fine approach to address these problems, which combines BEV (Bird's Eye View) feature extraction, coarse-grained matching and fine-grained verification. In the coarse stage, our approach utilizes an attention-guided network to generate attention-guided descriptors. We then employ a fast affinity-based candidate selection process to identify the Top-\textit{K} most similar candidates. In the fine stage, we estimate pairwise overlap among the narrowed-down place candidates to determine the final match. Experimental results on the KITTI and KITTI-360 datasets demonstrate that our approach outperforms state-of-the-art methods. The code will be released publicly soon.
\end{abstract}


\section{INTRODUCTION}
Place recognition is a crucial task in mobile robots and autonomous driving, as it enables the recognition of previously visited places and provides a basis for loop closure detection. It plays a vital role in Simultaneous Localization and Mapping (SLAM) by identifying loop closures to correct drift and tracking errors. The LiDAR sensor has gained popularity for place recognition due to its robustness to illumination and weather changes and its wide field of view.

Recent LiDAR-based place recognition methods focus on describing the LiDAR point cloud discriminatively using various representation forms, including 3D point clouds\cite{uy2018cvpr, vid2022icra, komorowski2022icpr, komo2021wacv}, segments\cite{kong2020iros, vid2021icra}, range image views\cite{ma2022ral, ma2022tie, yin2020iros, zhao2023ral,wang2020iros}, and bird's eye views\cite{kim2018iros, kim2021tro,wang2020icra, li2021iros, xu2021ral, luo2021ral, jiang2023icra, luo2023}. These methods employ hand-crafted or learning-based techniques to generate efficient local or global descriptors. However, compressing the LiDAR data inevitably limits the representation ability of descriptors. Besides, totally different places may have similar descriptions, which can result in failures for loop closure detection. On the other hand, pairwise similarity-based place recognition methods have gained attention\cite{chen2020rss, li2022ral, li2023icra}. These methods compare pairs of point clouds using a network to predict their similarity score. Pairwise comparison fully utilizes the information between two point clouds, improving loop closure detection performance. However, pairwise similarity-based methods require exhaustive searches to detect loops, making them time-consuming.

 \begin{figure}[t]
	\centering
            \begin{minipage}{\linewidth}
            \centering
            \includegraphics[width=\linewidth]{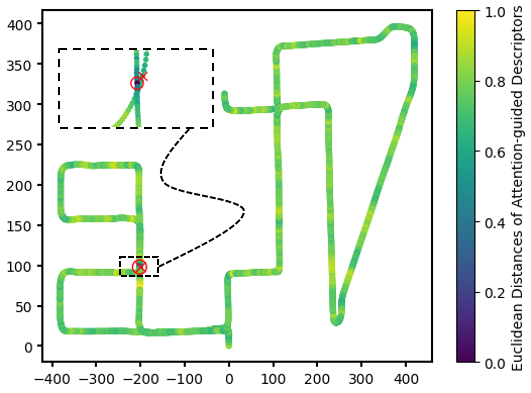} \\
            \end{minipage}

	\caption{The visualization of results. The trajectory represents sequence 08 of the KITTI dataset\cite{geiger2013ijrr}, and the color indicates the Euclidean distance of the descriptors to the query. The circle mark represents a query place, and the cross mark represents the matching place found by our approach.
 }
	\label{pic: illustration}
\end{figure}

To address these challenges, we propose a novel approach that efficiently combines the advantages of descriptors and pairwise similarity-based methods for coarse-to-fine place recognition. Our approach comprises three parts: BEV (Bird's Eye View) feature extraction, coarse-grained matching, and fine-grained verification. We first voxelize the input point cloud to create a multi-layer BEV representation. Next, we extract BEV features in 2D space and fully utilize them in our coarse-to-fine approach. In the coarse stage, an attention-guided network generates 1D global descriptor vectors from encoded BEV features. We then perform affinity-based candidate selection to obtain the Top-\textit{K} place candidates. Finally, in the fine stage, the candidates are further validated by estimating pairwise overlap using an overlap estimation network. The candidate with the highest overlap score is considered the final match. This combination of efficiency and accuracy highlights the advantages of our proposed method. Our approach effectively reduces the exhaustive search required for pairwise overlap estimation, improving the overall efficiency. Consequently, our method demonstrates superior performance in determining the matching place compared to existing methods. Fig.\ref{pic: illustration} provides an illustration of our proposed approach.

The main contributions of this paper are as follows:

\begin{itemize}
    \item An attention-guided global descriptor for effective place recognition.
    \item A coarse-to-fine place recognition approach that effectively uses BEV features to enhance efficiency while maintaining a high level of place recognition ability.
    \item Comprehensive experiments and detailed ablation studies on the KITTI\cite{geiger2013ijrr} and KITTI-360\cite{liao2022tpami} datasets to validate the effectiveness of the proposed approach.
\end{itemize}

\section{RELATED WORK}
Place recognition has been an active area of research for many years, especially in visual place recognition (VPR), where many works have been proposed\cite{chen2017icra, gehrig2017icra, schubert2020icra, camara2020icra, zhang2021icra}. In this paper, we focus on LiDAR-based place recognition (LPR).

\textbf{Description-based Methods.} Description-based methods aim to provide discriminative descriptions of LiDAR point clouds for place recognition. Traditional methods often use hand-crafted features\cite{he2016iros, kim2018iros,kim2021tro, wang2020icra, li2021iros}, while more recent approaches leverage the neural networks to achieve powerful feature representation. Some description-based methods take raw point clouds or voxelized and segmented forms as input to the network model\cite{uy2018cvpr, kong2020iros, vid2021icra, vid2022icra, komorowski2022icpr, komo2021wacv}. For example, PointNetVLAD\cite{uy2018cvpr} combines PointNet\cite{qi2017cvpr} and NetVLAD\cite{arandjelovic2016cvpr} to generate global descriptors in an end-to-end manner. Locus\cite{vid2021icra} leverages topological and temporal information and aggregates multi-level features via second-order pooling (O2P). LoGG3D-Net\cite{vid2022icra} uses O2P followed by Eigen-value Power Normalization to obtain a global descriptor. To improve efficiency, some methods use projections of LiDAR data. M2DP\cite{he2016iros}, for example, projects a 3D point cloud onto multiple 2D planes to form a descriptor. Range images are also widely used\cite{chen2020rss, ma2022ral, ma2022tie, yin2020iros, zhao2023ral,wang2020iros}, which are consistent with the characteristics of LiDAR sensors. Thanks to the property of range images, it is easy to obtain orientation invariance theoretically. However, small translations may affect the range image due to perspective transformation. Some methods use bird's eye view forms of data as input\cite{kim2018iros,kim2021tro, wang2020icra, li2021iros, xu2021ral, luo2021ral, jiang2023icra, luo2023}, as Cartesian coordinate representation is more consistent with reality. Some methods use a single-layer BEV, such as Scan Context\cite{kim2018iros}, Intensity Scan Context\cite{wang2020icra}. Xu et al.\cite{xu2021ral} summarize several BEV representations, including multi-layer occupied BEV, multi-layer density BEV, and single-layer height BEV. The multi-layer BEV can reduce the loss of height information. Furthermore, CVTNet\cite{ma2023tii} fuses range image views and bird's eye views of LiDAR data and achieves good performance.

\textbf{Pairwise Similarity-based Methods.} Pairwise similarity-based methods focus on comparing two point clouds and scoring their similarity using specialized network structures\cite{kong2020iros, chen2020rss, li2022ral, li2023icra}. OverlapNet\cite{chen2020rss}, for example, exploits multiple cues, including depth, normal, intensity, and semantic class probability information to predict the overlap score and treats overlap estimation as a regression problem. RINet\cite{li2022ral} also takes a similar approach to predict similarity. It combines semantic and geometric features to predict the similarity of descriptor pairs. These methods perform pairwise similarity comparison implicitly. In contrast, our previous work\cite{li2023icra} proposed an intuitive approach. It regards overlap prediction as a classification problem of each bin in the bird's eye view and achieves state-of-the-art performance in loop closure detection. However, with the promotion of place recognition ability, the time cost increases significantly to perform exhaustive pairwise overlap estimation.

Cop et al.\cite{cop2018icra} design an intensity-based local descriptor DELIGHT and propose a two-stage approach to perform place recognition. In the first stage, they search for the \textit{N} most similar DELIGHT descriptors and subsequently perform geometrical recognition on the identified place candidates. Inspired by their work, we propose a coarse-to-fine approach that combines the advantages of description-based methods and pairwise similarity-based methods. We fully leverage the extracted BEV features to implement the coarse-to-fine process efficiently. In the coarse stage, these features are employed to generate attention-guided descriptors and enable fast affinity-based candidate selection. In the fine stage, we utilize the selected corresponding features to perform pairwise overlap estimation among the narrowed-down place candidates. By doing so, we efficiently detect loop closures and achieve leading performance compared to other state-of-the-art methods.

 \begin{figure*}[t]
    \centering
    \includegraphics[width=2\columnwidth]{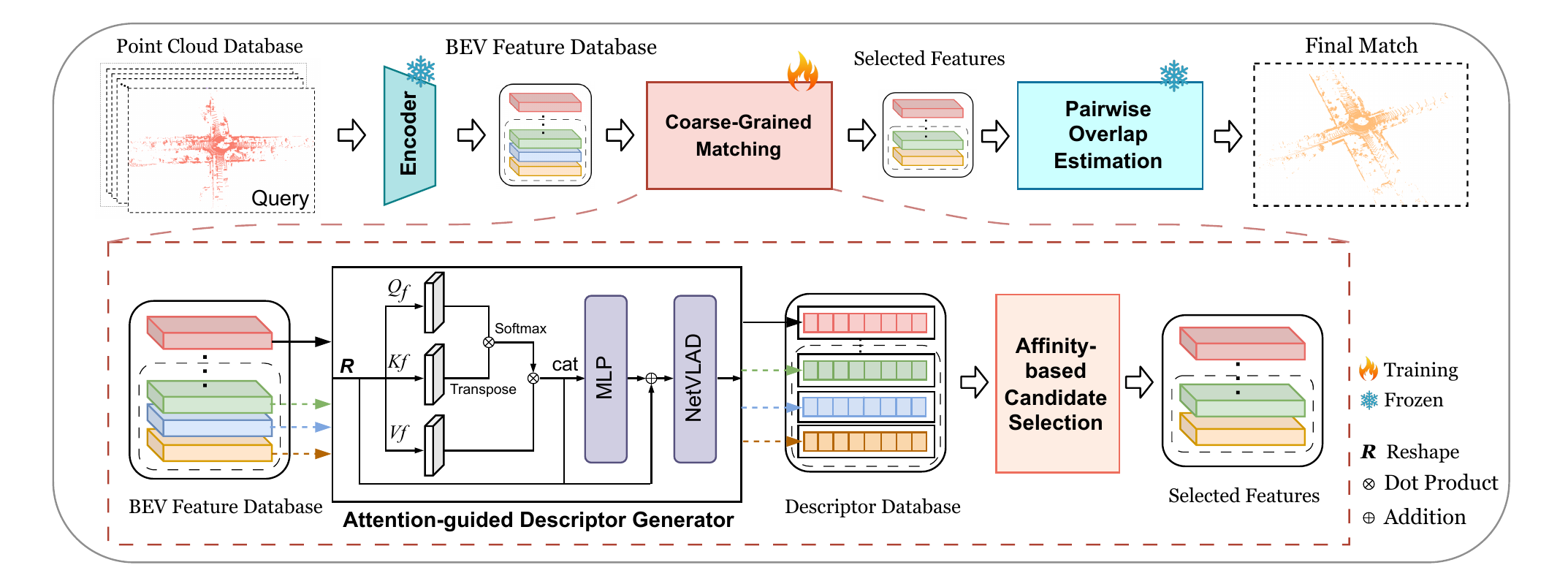}
    \caption{The pipeline of our approach. The point clouds are first converted to voxel representations and fed into the encoder network to extract BEV features. Then, these features are used to generate global descriptors. In the coarse phase, the Top-\textit{K} place candidates are selected by affinity-based selection. Then, in the fine phase, the corresponding BEV features are used pairwise to estimate the overlap region between the query scan and the candidates to find the final match.}\vspace{-6mm}
    \label{pic:overview}
 \end{figure*}
 
\section{METHODOLOGY}
Our approach employs a coarse-to-fine methodology, which consists of three primary modules: BEV feature extraction, coarse-grained matching using attention-guided descriptors, and fine-grained verification through pairwise overlap estimation. Firstly, the encoder extracts BEV features from multi-layer BEVs in 2D space. The features are subsequently used to generate an attention-guided global descriptor. Candidate scans are then rapidly retrieved through affinity-based candidate selection. Finally, we perform pairwise overlap estimation among the narrowed-down place candidates to find the final match. Additional design details, such as the loss function, are also discussed. An overview of our approach is illustrated in Fig.~\ref{pic:overview}.

\subsection{BEV Feature Extraction} \label{sec_bev}
The BEV feature extraction process follows our previous work \cite{li2023icra}. This process begins by voxelizing the input point cloud to create a voxel grid, where each voxel is assigned a value of 0 or 1 depending on whether any points are present within it. This transformation results in the creation of multi-layer BEVs, denoted as $B\in\mathbb{R}^{H_B\times W_B\times C_B}$. Each layer is regarded as a channel during feature extraction. The feature encoder module is a 2D sparse convolution network with residual blocks. The resulting BEV feature volume is denoted as $f \in H_f \times W_f \times C_f$. $f$ is stored in the database $\text{DB}_f$ for later global descriptor generation and pairwise overlap estimation. The BEV features are particularly advantageous for detecting closed loops because they are insensitive to minor translations due to the voxelization operation and exhibit translation invariance due to the convolution operation.

\subsection{Coarse-Grained Matching} \label{sec_coarse}

\textbf{Attention-guided Descriptor Generator.} The attention-guided descriptor generation network consists of a self-attention module, a NetVLAD layer, and a fully connected layer. Since the BEV feature only contains information in a limited local neighborhood, it results in insufficient learning of patterns from the local feature volume for the global descriptor. Therefore, we utilize a self-attention module\cite{zhang2019icml} to gain contextual information and further contribute to better performance in global descriptors generation.

\textit{1) Self-Attention on BEV feature:} As the input of the attention module is required to be a sequence, we first reshape the BEV feature $f$ into $f \in \mathbb{R}^{C_f \times L_f}$, where $L_f = H_f \times W_f$. Then $f$ is transformed into three different feature spaces: query feature space $Q_f = W_q f$, key feature space $K_f = W_k f$ and value feature space $V_f = W_v f$, where $W_q, W_k, W_v\in\mathbb{R}^{C_f \times C_f}$ and $Q_f, K_f, V_f\in\mathbb{R}^{C_f \times L_f}$. The generated contextual information $A\in\mathbb{R}^{C_f \times L_f}$ is as follows:

\begin{equation}
    A = \text{softmax}( \frac{Q_fK_f^T}{\sqrt{d_k}} )V_f,
\end{equation}

where $d_k = C_f$. Then the contextual information $A$ and original BEV feature $f$ are used to yield a new feature $f'$:

\begin{equation}
    f' = f + \text{MLP}(\text{cat}(f, A)),
\end{equation}
where $\text{MLP}(\cdot)$ denotes a three-layer fully connected network, $\text{cat}(\cdot,\cdot)$ means concatenation.

The self-attention module allows each local descriptor in the original feature $f$ to gather contextual information from other local descriptors. By enlarging the receptive field and capturing spatial relations between local descriptors, the attention mechanism enhances the performance of feature extraction in the BEV representation. Furthermore, compared to convolutional networks, self-attention has been shown to improve the detection of reverse loops, as confirmed by our experiments in Section \ref{sec_as}.

\textit{2) Global descriptor generator:} Then the feature $f'$ is fed into the NetVLAD layer to generate a global descriptor. NetVLAD is a neural network layer that aggregates local features into a compact and discriminative global descriptor by learning a set of cluster centers and soft assignments. The NetVLAD layer aggregates local descriptors $\textbf{x}_i$ within the feature $f'$ by summing the residuals between each descriptor and cluster centers, which are weighted by soft-assignment of descriptors to multiple clusters. The output VLAD representation is as follows: 

\begin{equation}
    V(j, k) = \sum_{i=1}^N \overline{a}_k(\textbf{x}_i) (x_i(j)-c_k(j)),
\end{equation}
where $\overline{a}_k$ denotes the soft-assignment, $\textbf{x}_i$ is the $i^{th}$ descriptor in $f'$, and $x_i(j)$ is the $j^{th}$ element of the $i^{th}$ descriptor, $c_k$ is the $k^{th}$ cluster center.

Finally, a fully connected layer is utilized to reduce the dimension of the VLAD representation and computational costs, which follows \cite{uy2018cvpr}. Therefore, we obtain the attention-guided global descriptor vector $v$.

\textbf{Affinity-based Candidate Selection.} Firstly, we build up a descriptor database $\text{DB}_v$ from $\text{DB}_f$ so that every scan has a discriminative and independent description. Next, we use affinity-based candidate selection to identify loop candidates from the rest in $\text{DB}_v$. To assess the affinity between descriptor vectors, we adopt the Euclidean distance, known for its simplicity and efficiency. The affinity is calculated as follows:

\begin{equation}
    Af_{PQ} =   \left\|v_P - v_Q\right\|_2,
\end{equation}
where $\left\| \cdot \right\|_2$ denotes the $L_2$ normalization. Using this simple calculation, we can efficiently select the Top-\textit{K} candidates with the \textit{K} closest Euclidean distances, significantly reducing the range for subsequent pairwise overlap estimation.

\subsection{Fine-Grained Verification} \label{sec_fine}
\begin{figure}[t]
	\centering
	\subfigure[]{
            \begin{minipage}[c]{0.3\linewidth}
            \centering
            \includegraphics[width=\linewidth]{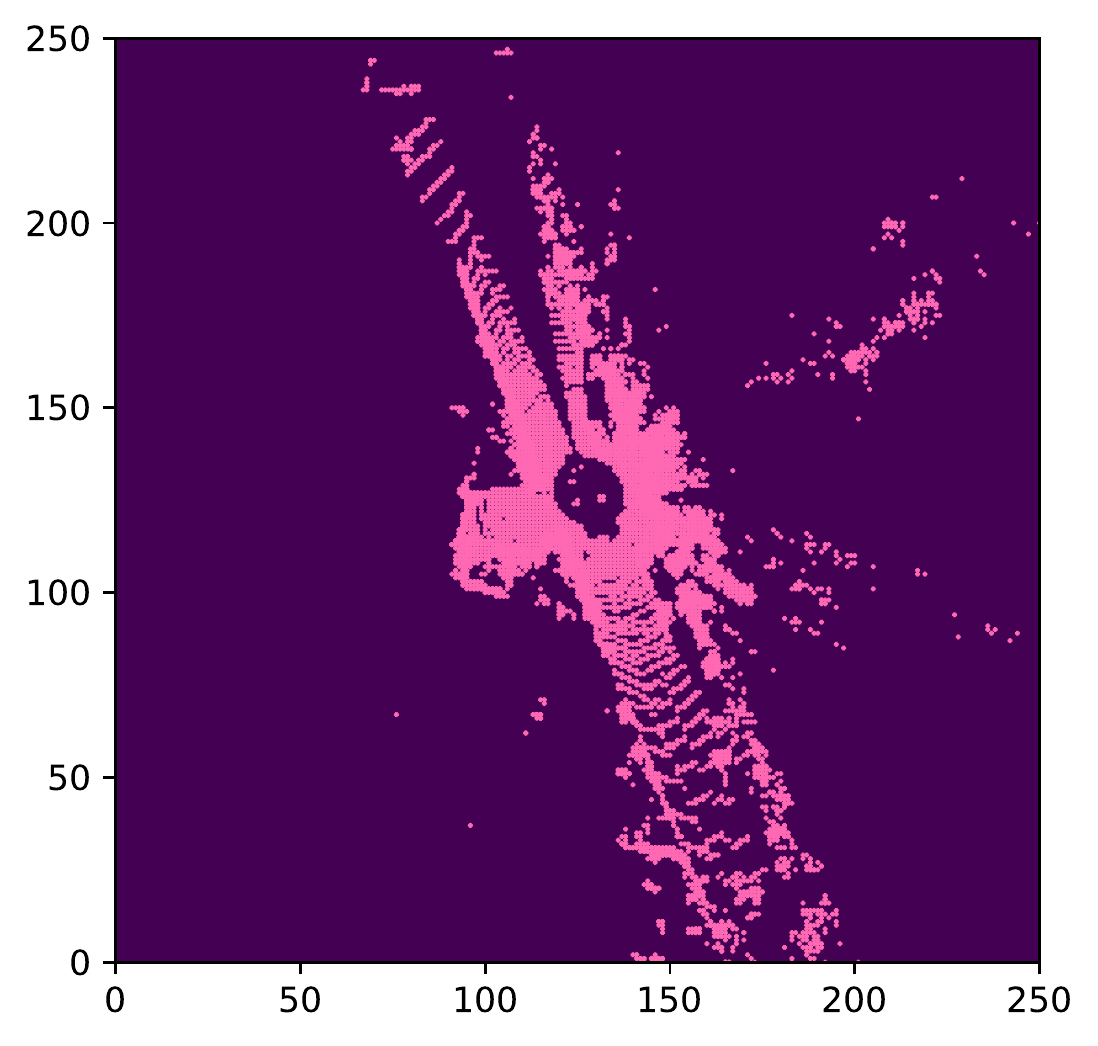}
            \includegraphics[width=\linewidth]{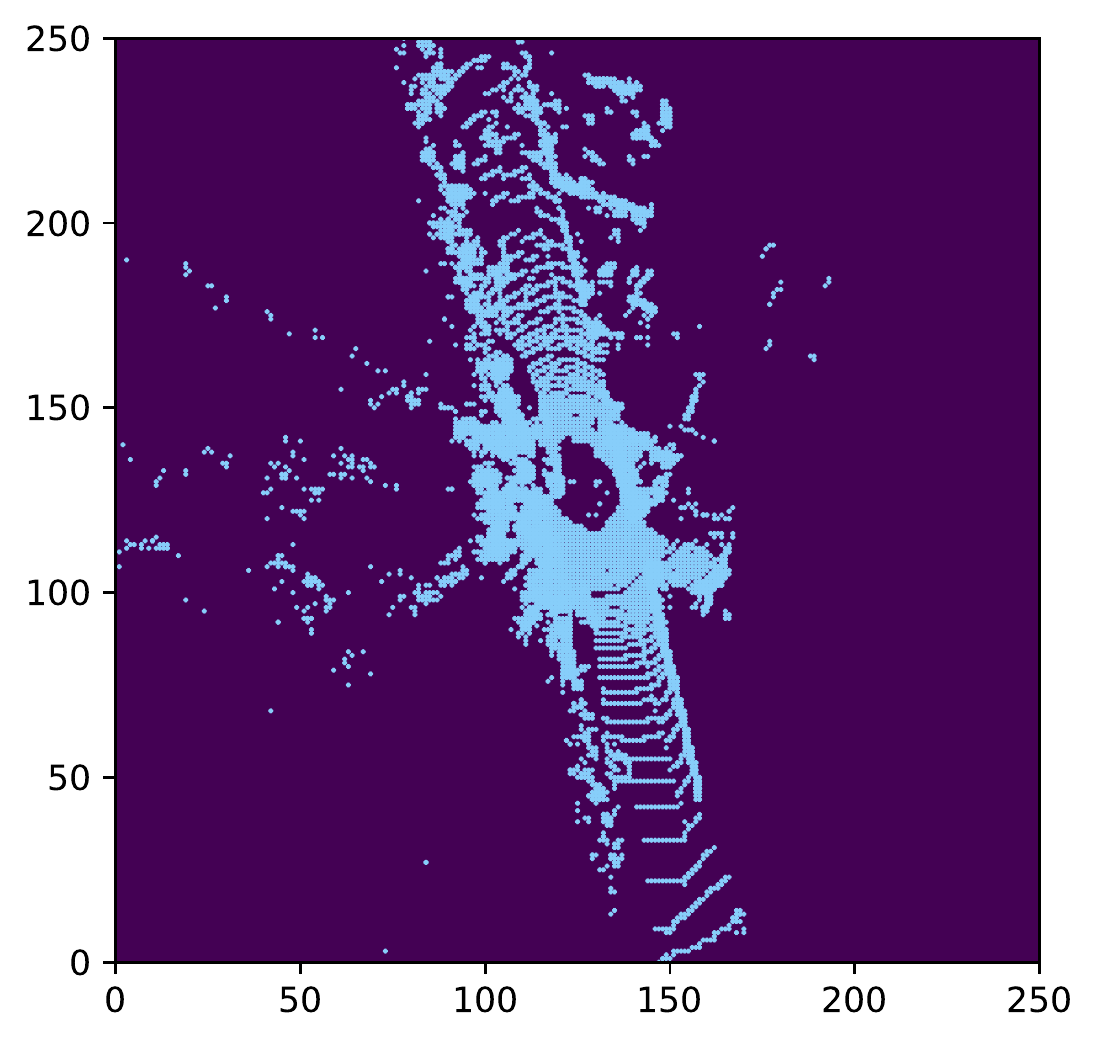}
            \end{minipage}
        }
	\subfigure[]{
            \begin{minipage}[c]{0.29\linewidth}
            \centering
            \includegraphics[width=\linewidth]{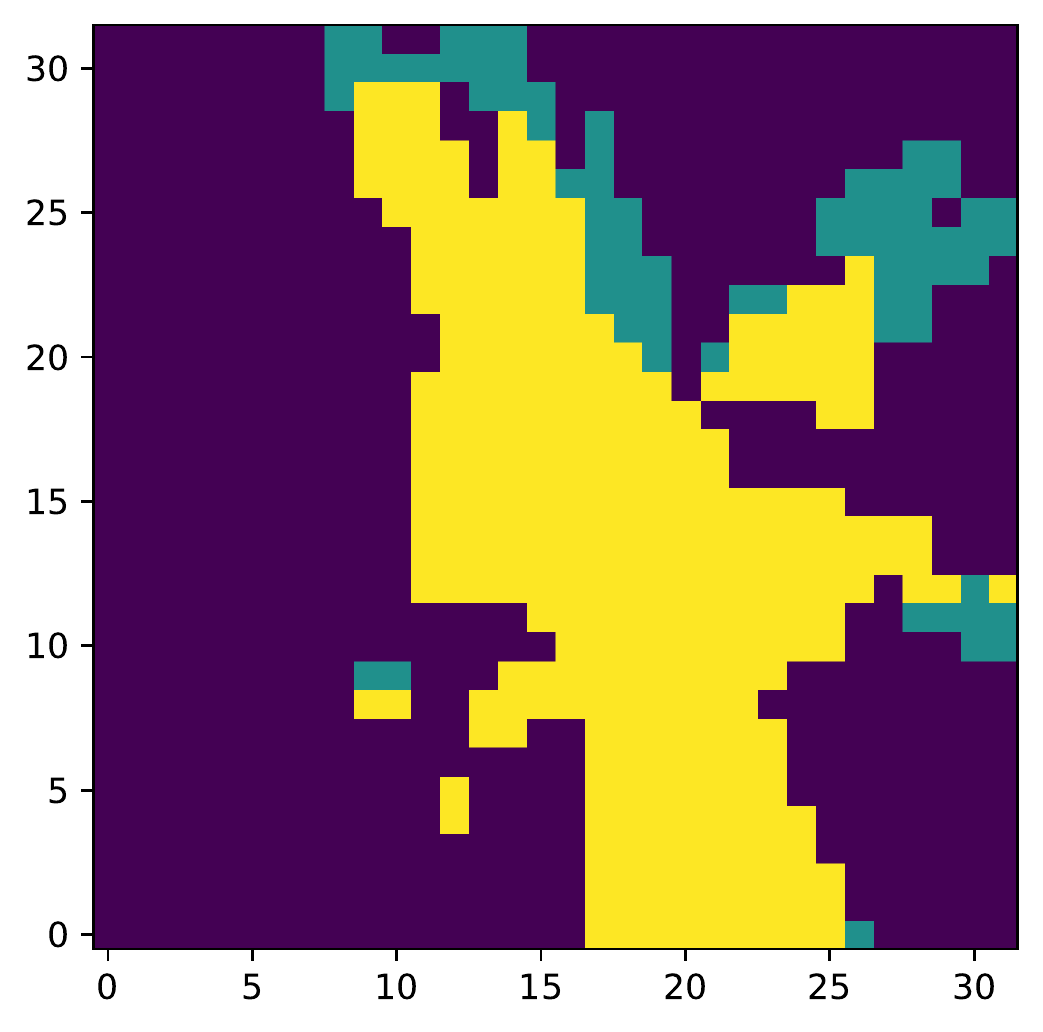}
            \includegraphics[width=\linewidth]{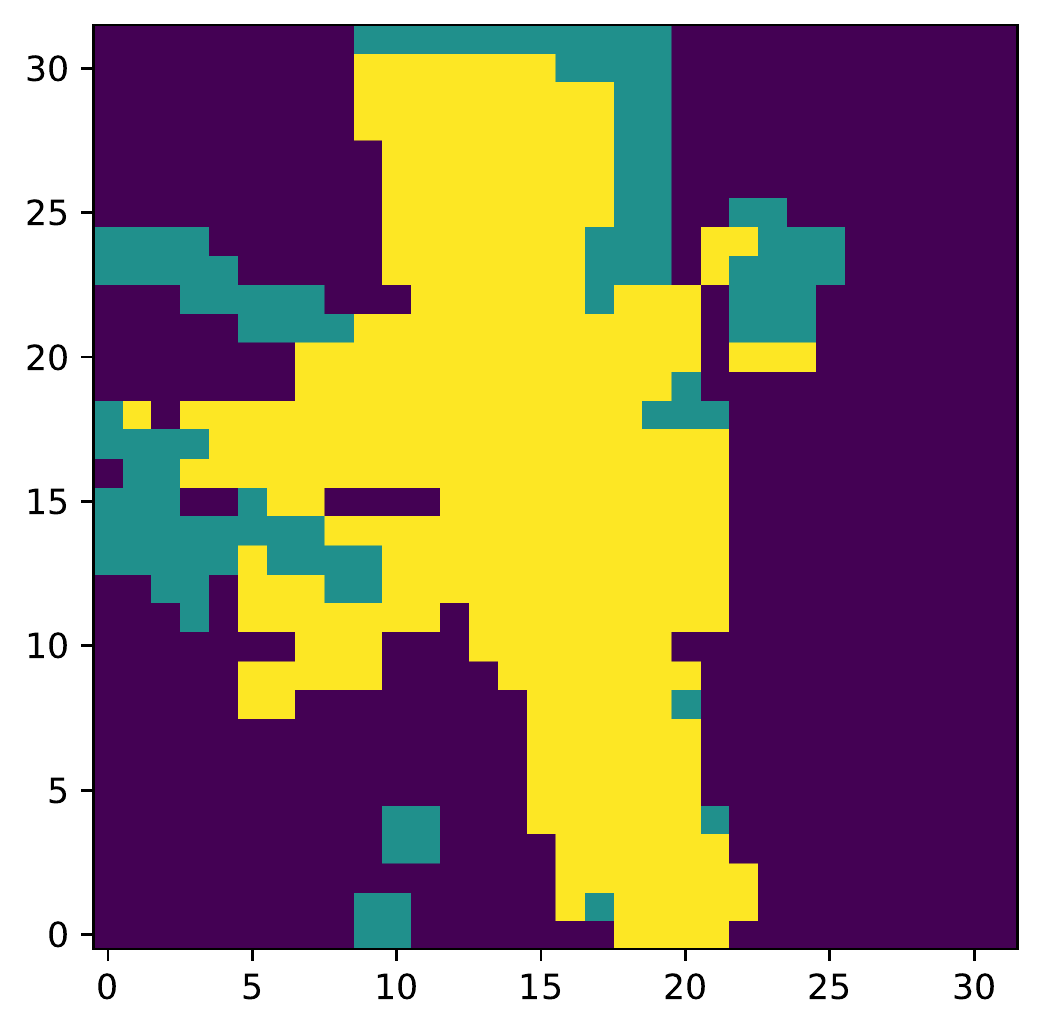}
            \end{minipage}
        }
	\subfigure[]{
            \begin{minipage}[c]{0.29\linewidth}
            \centering
            \includegraphics[width=\linewidth]{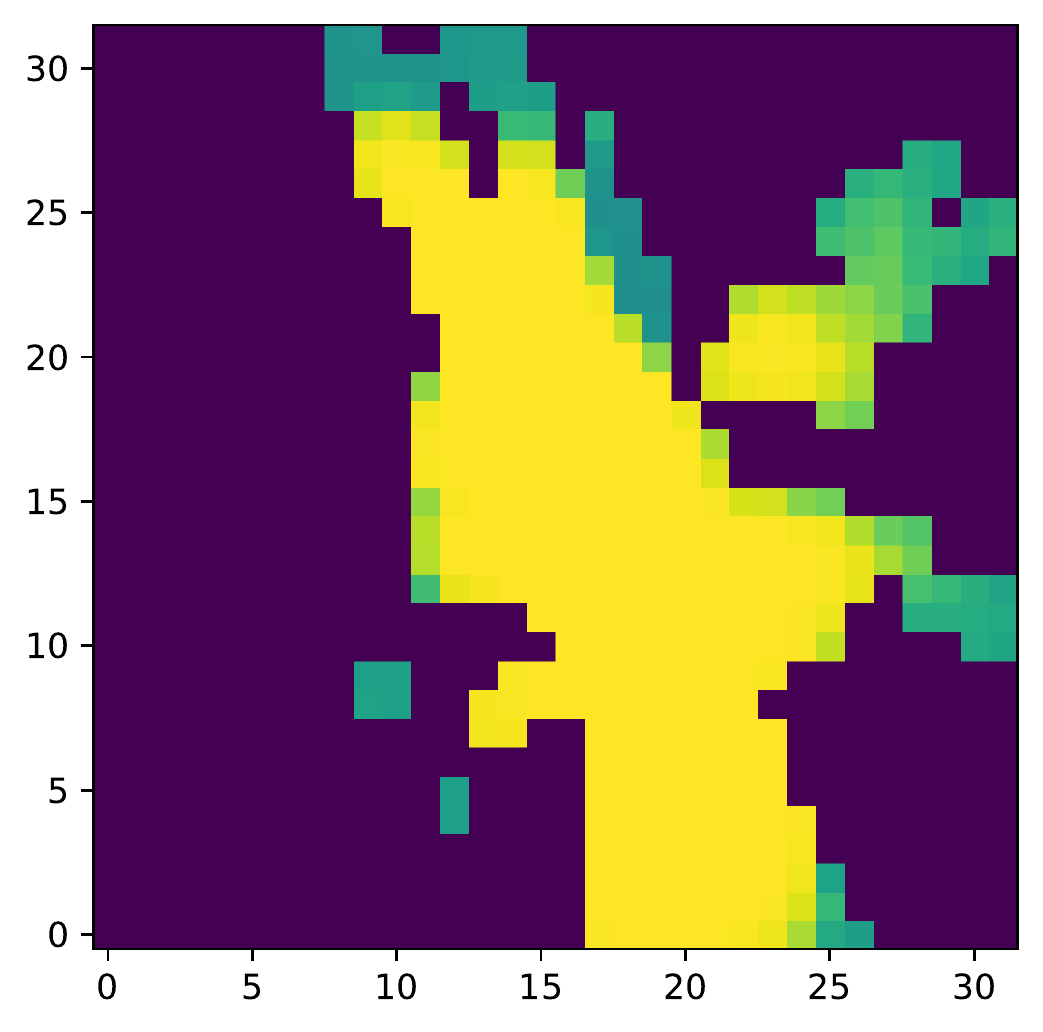}
            \includegraphics[width=\linewidth]{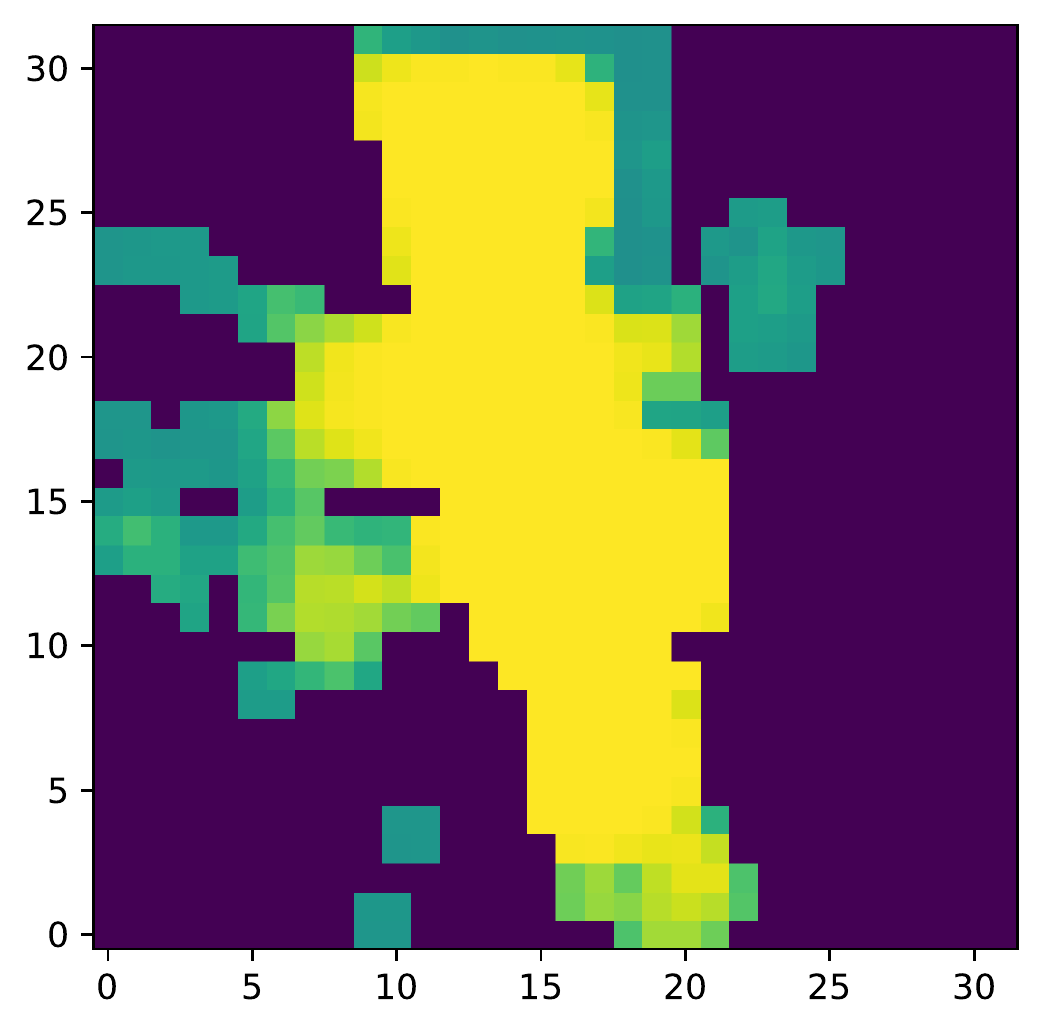}
            \end{minipage}
        }
	\caption{The figure above depicts the process of overlap estimation, where the top row corresponds to the query scan, and the bottom row corresponds to the candidate. Panel (a) shows the input pairwise point clouds. The yellow region in panel (b) represents the ground truth overlap region and panel (c) shows the predicted overlap region.}
	\label{pic:overlap}
\end{figure}

In the fine stage, we determine the most similar place among the Top-\textit{K} candidates by performing pairwise overlap estimation. The overlap estimation module\cite{li2023icra} comprises a cross-attention module and a classification head. The cross-attention module effectively enhances feature representation by facilitating information interaction and contextual aggregation. The overlap estimation process is illustrated in Fig.\ref{pic:overlap}.

The cross-attention module fuses the relevant feature maps $r_P$ and $r_Q$ from pairwise BEV features $f_P$ and $f_Q$ as follows:

\begin{align}
\begin{split}
    r_P &= f_P + \text{MLP}(\text{cat}(f_P, \text{att}(f_P, f_Q, f_Q))) \\
    r_Q &= f_Q + \text{MLP}(\text{cat}(f_Q, \text{att}(f_Q, f_P, f_P))),
\end{split}
\end{align}
where $\text{MLP}(\cdot)$ denotes a fully connected network, $\text{cat}(\cdot,\cdot)$ means concatenation and $\text{att}(\cdot,\cdot,\cdot)$ is the attention model.

The overlap estimation is regarded as a binary classification problem. The classification head consists of two $3\times3$ convolutional layers and a sigmoid layer with ReLU as the activation function as follows:
\begin{equation}
\small
    \gamma_t = \text{Sigmoid}(\text{Conv}_{3\times3}(\text{ReLU}(\text{Conv}_{3\times3} (r_t)))), t\in\{P,Q\}.
\end{equation}

Thus, we can get the overlap score as follows:
\begin{equation}
    \tau = \frac{1}{2} \left(\frac{\sum \gamma_P}{N_P} + \frac{\sum \gamma_Q}{N_Q} \right),
\end{equation}
where $N_P$ and $N_Q$ denote the total number of non-zeros pixels of $r_P$ and $r_Q$.

By quantifying the overlapping area with the overlap score to evaluate the similarity of pairwise point clouds, we can get the candidate with the highest overlap score as the final match.

\subsection{Loss Function} \label{sec_loss}
In our approach, we only train the attention-guided descriptor generation network using lazy triplet loss\cite{uy2018cvpr}. The BEV feature extraction network and overlap estimation network use the pre-trained model of our previous work\cite{li2023icra}. The lazy triplet loss is defined as follows:

\begin{equation}
    \mathcal{L}=max([m+\delta_{pos_i}-\delta_{neg_j}]_+),
\end{equation}
where $[\cdot]_+$ represents the hinge loss, $m$ is a constant margin, and $\delta(\cdot)$ is the descriptor distances between the query and positive/negative samples. We consider point clouds that are less than $\sigma_{pos}$ meters away as positive samples and those more than $\sigma_{neg}$ meters away as negative samples.

\section{EXPERIMENTS}
To demonstrate the effectiveness and practicality of our proposed approach for LiDAR-based place recognition, we design a series of experiments. We evaluate our method on the widely-used KITTI and KITTI-360 datasets and compare its performance with other state-of-the-art methods. Furthermore, we conduct ablation studies to investigate the impact of the coarse matching stage on the final results. Additionally, runtime experiments are performed to evaluate the computational efficiency of our method.

\begin{figure*}[t]
    \centering
        \subfigure[00]{
        \centering
        \includegraphics[width=0.46\columnwidth]{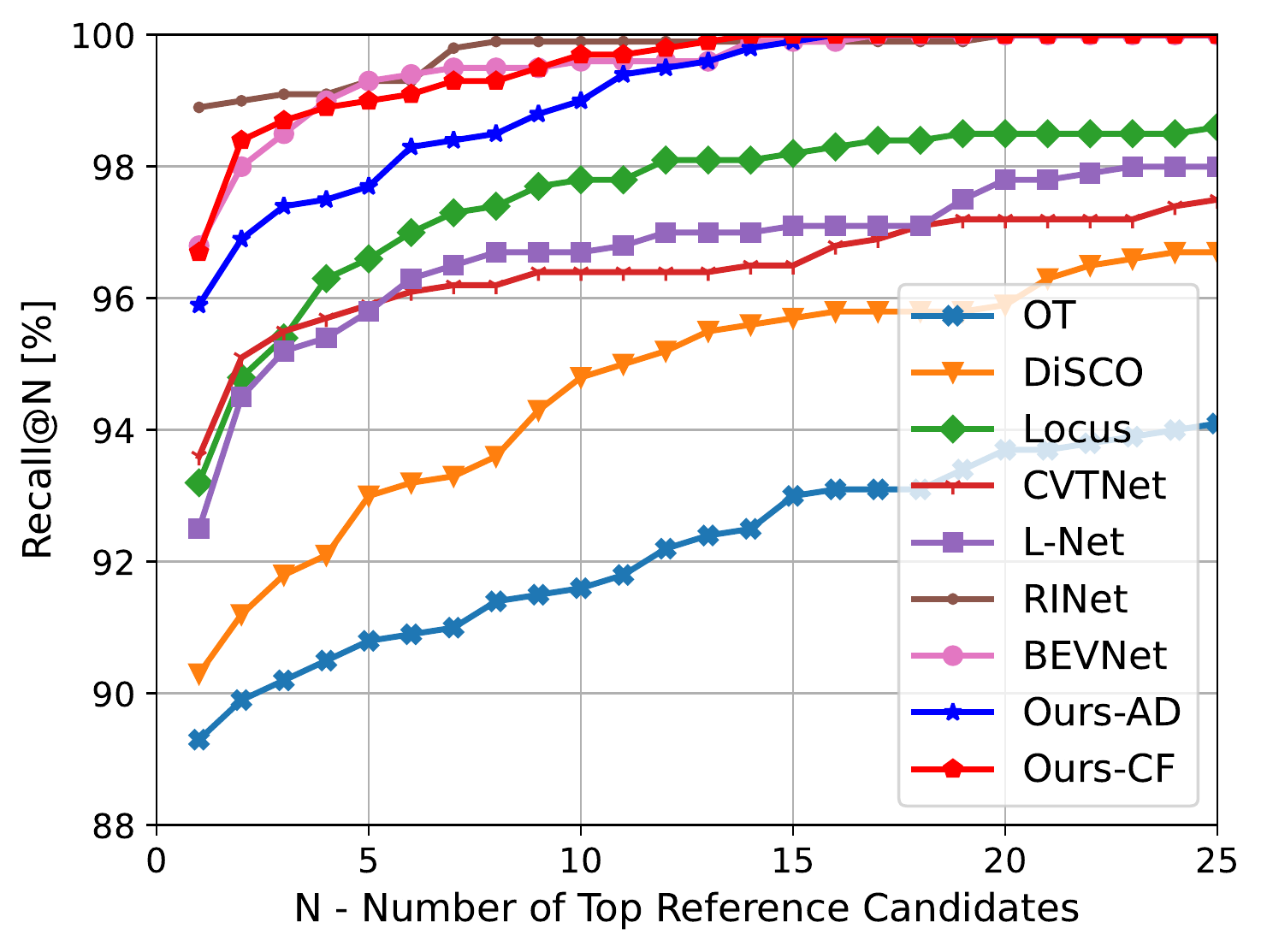}
        }
        \subfigure[02]{
        \centering
        \includegraphics[width=0.46\columnwidth]{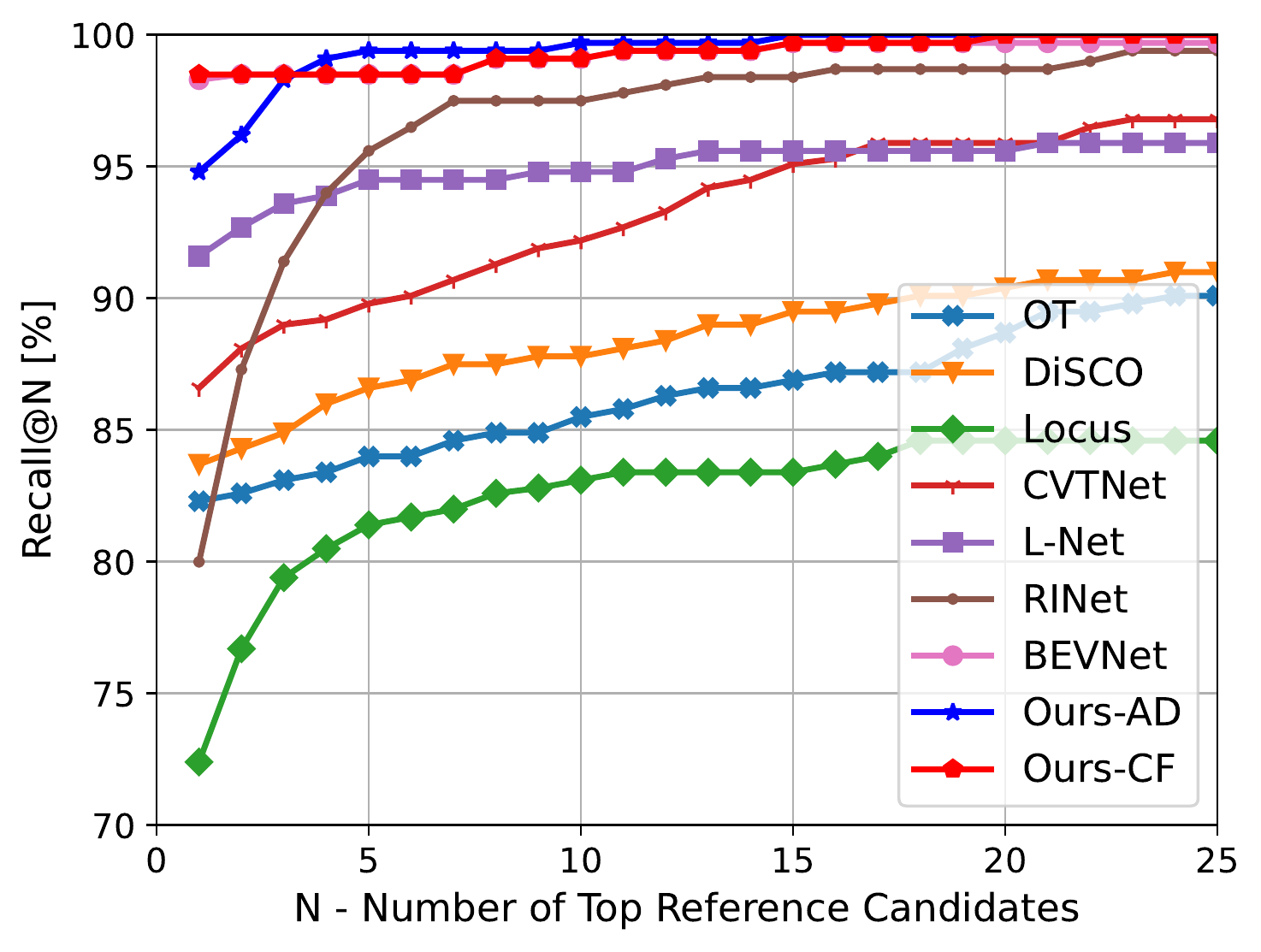}
        }
        \subfigure[05]{
        \centering
        \includegraphics[width=0.46\columnwidth]{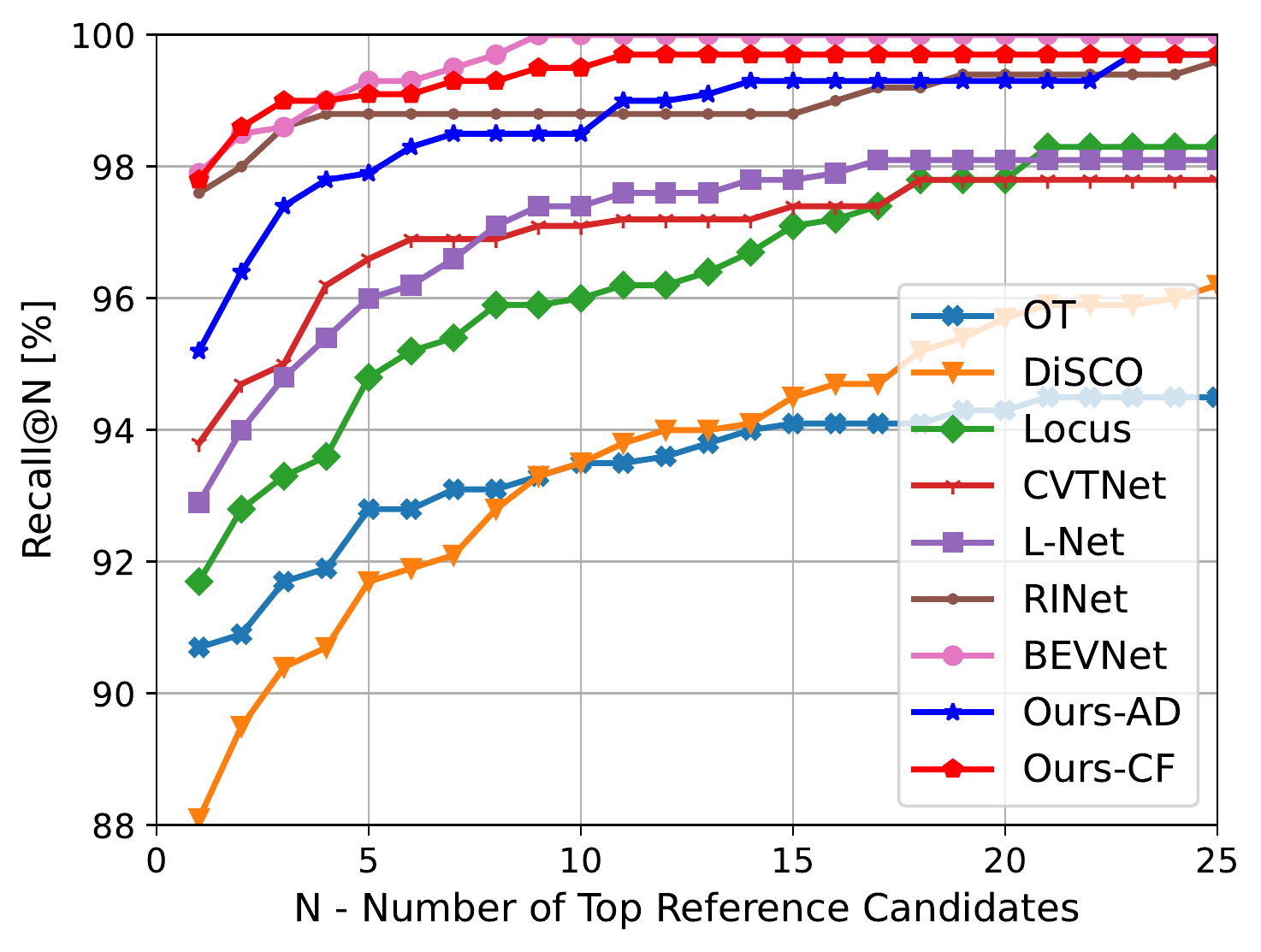}
        }
        \subfigure[0002]{
        \centering
        \includegraphics[width=0.46\columnwidth]{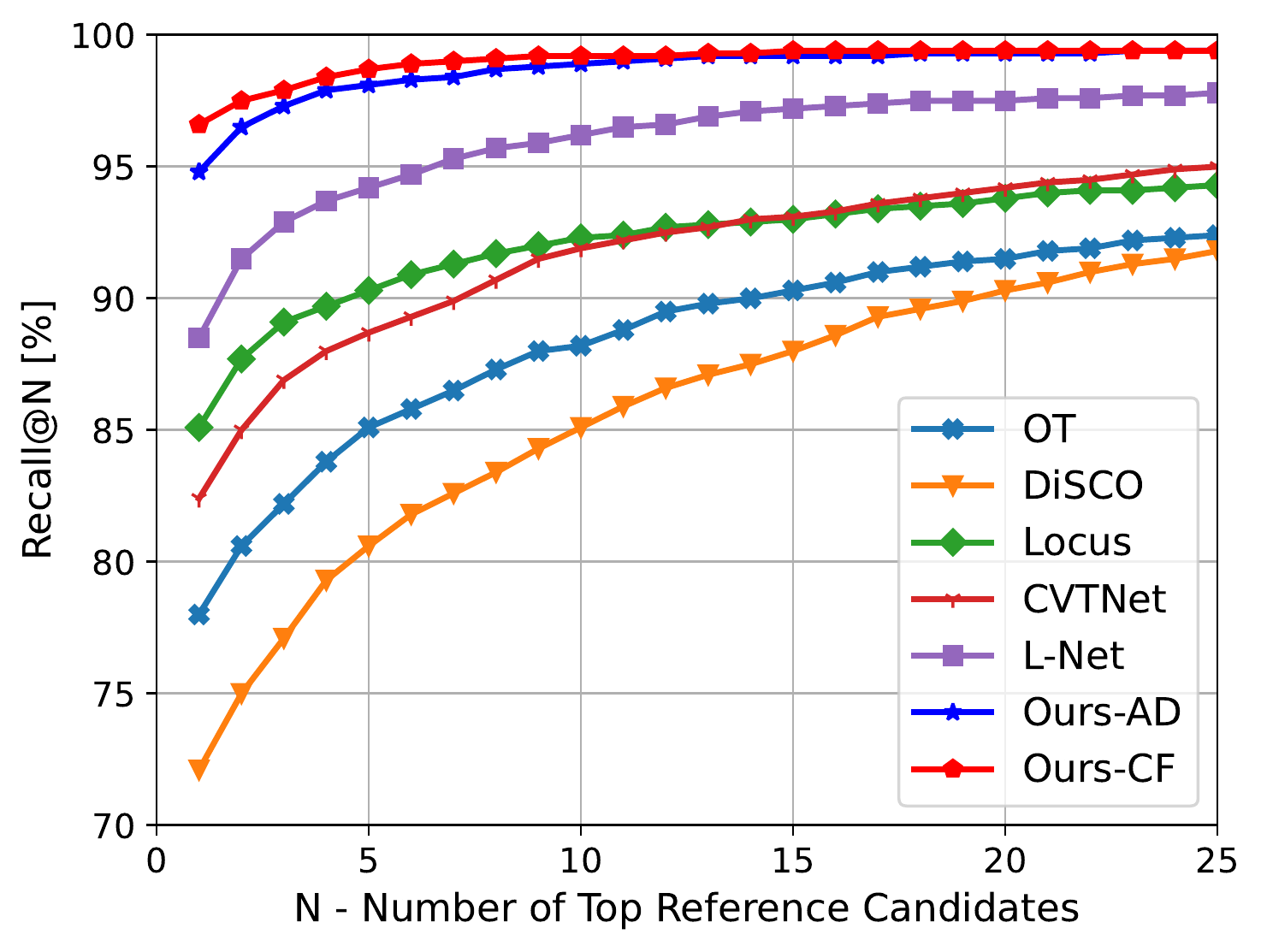}
        }
        \vspace{-3mm}

        \subfigure[06]{
        \centering
        \includegraphics[width=0.46\columnwidth]{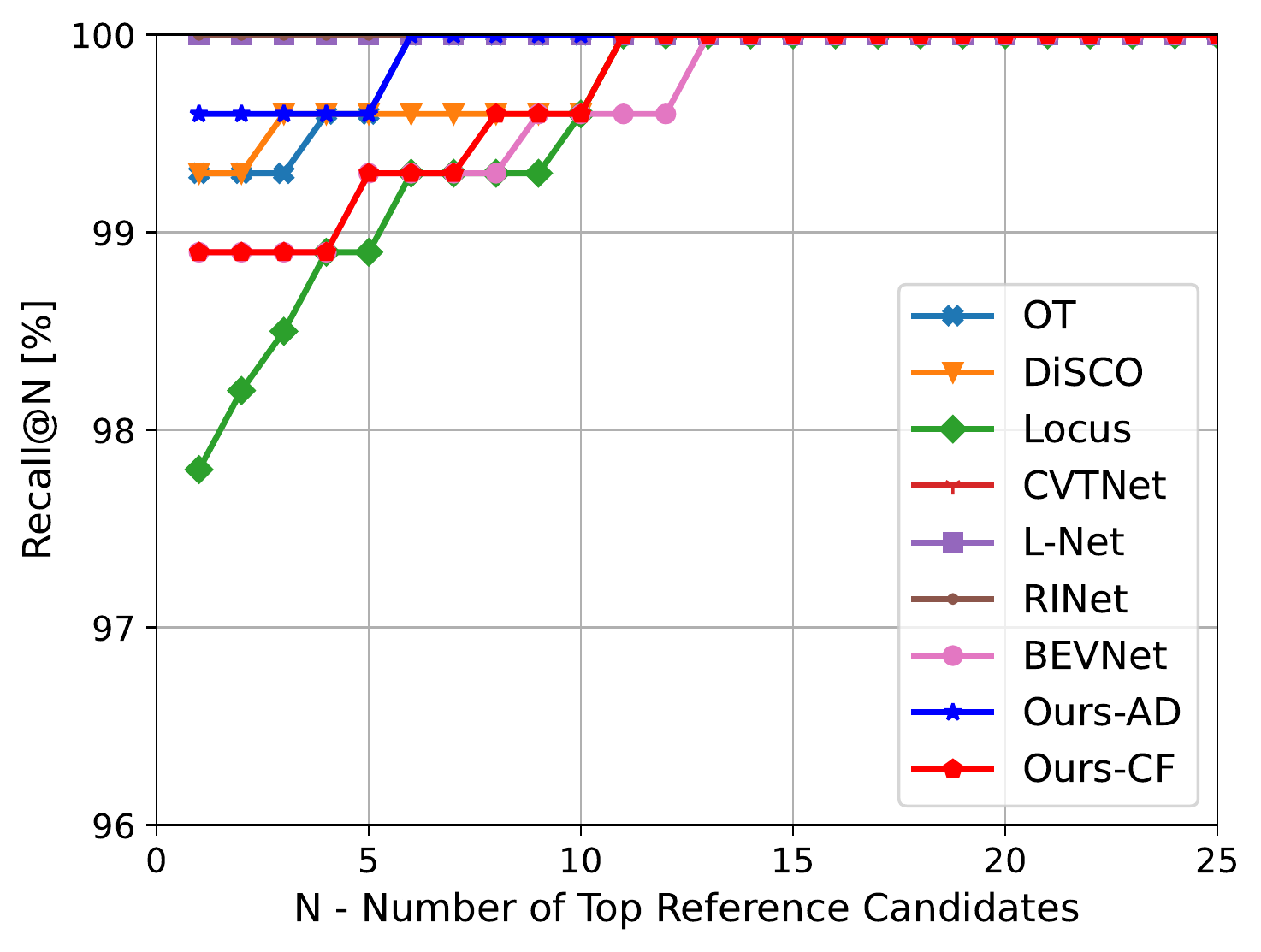}
        }
        \subfigure[07]{
        \centering
        \includegraphics[width=0.46\columnwidth]{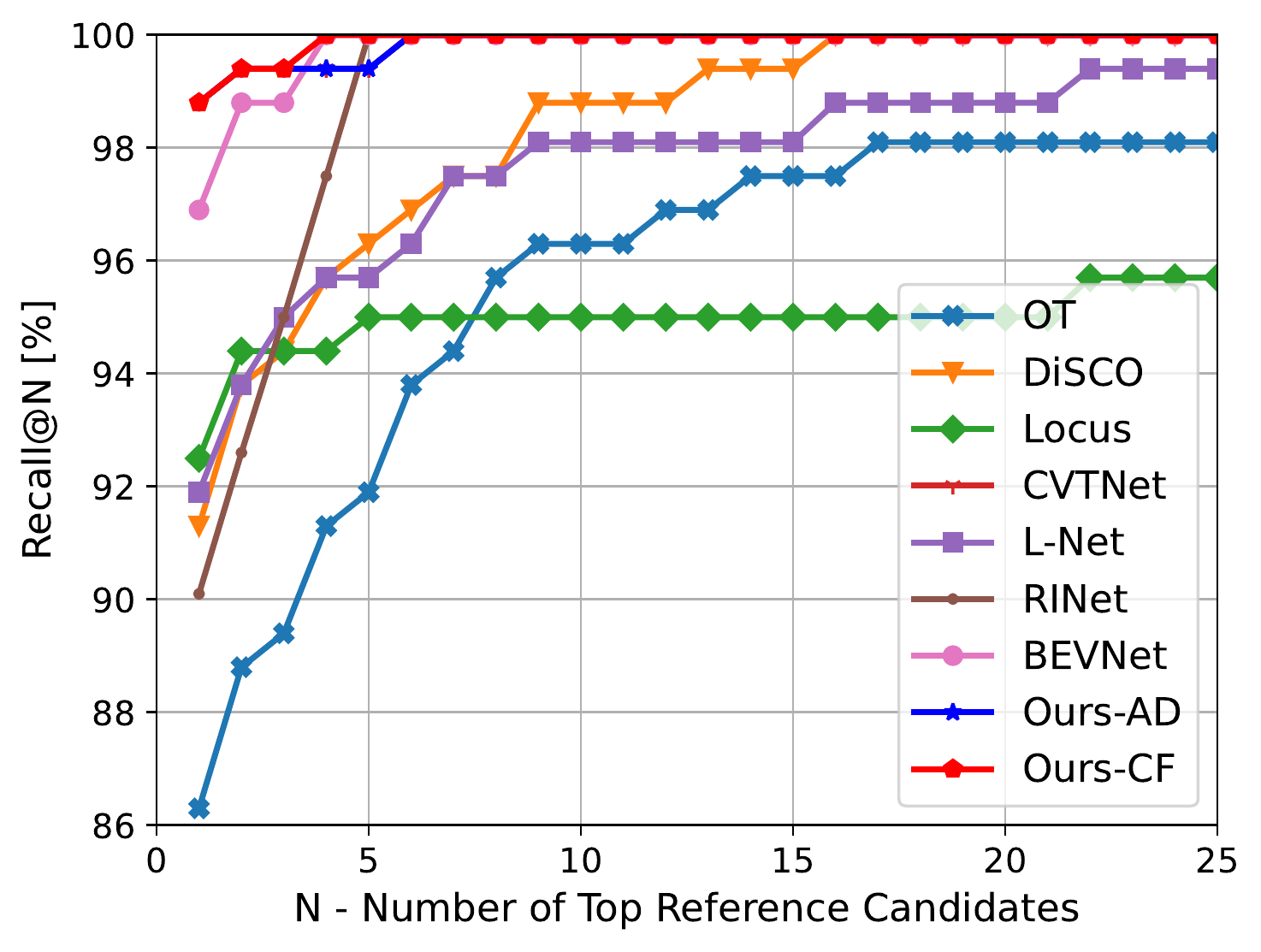}
        }
        \subfigure[08]{
        \centering
        \includegraphics[width=0.46\columnwidth]{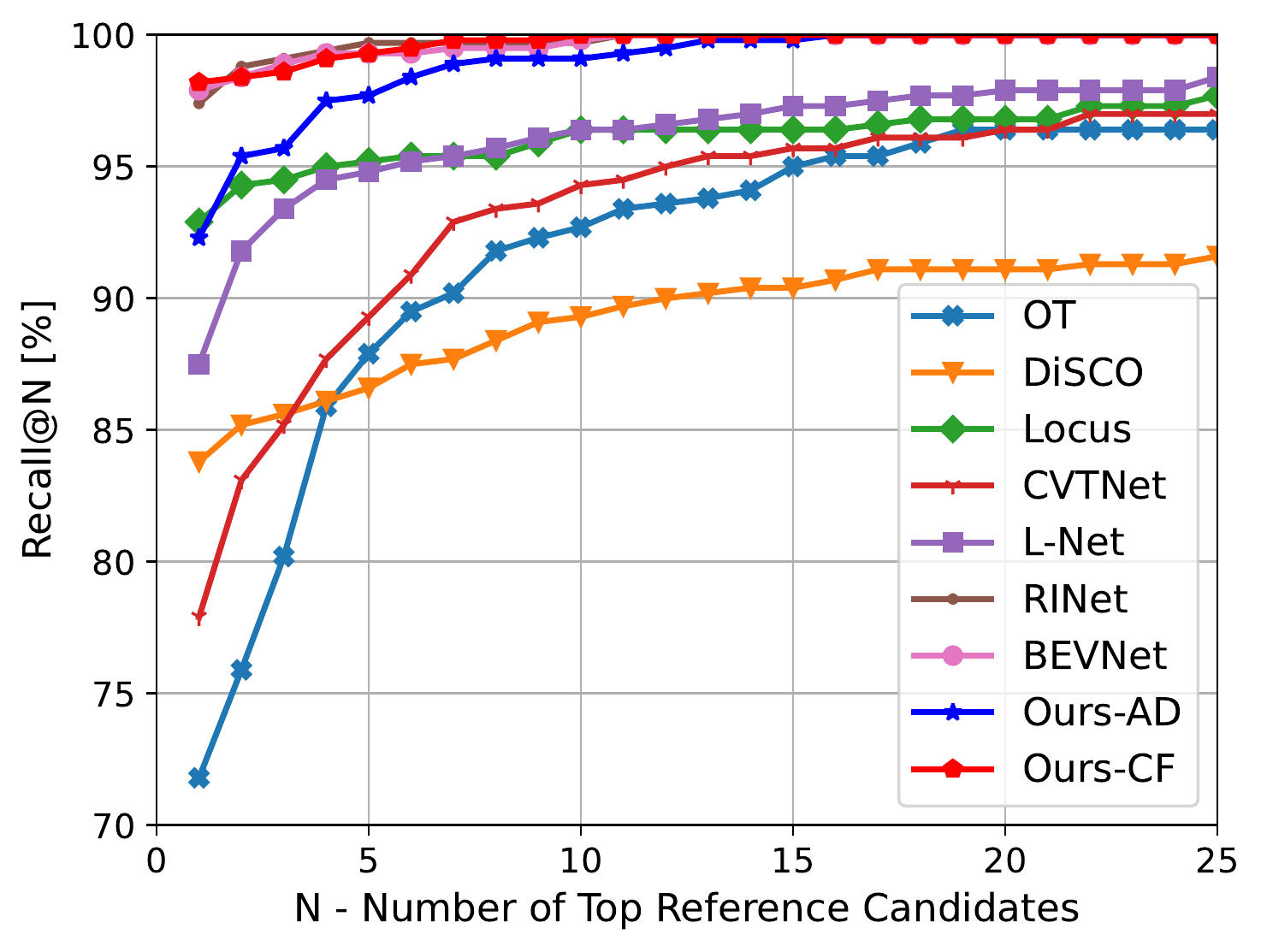}
        }
        \subfigure[0009]{
        \centering
        \includegraphics[width=0.46\columnwidth]{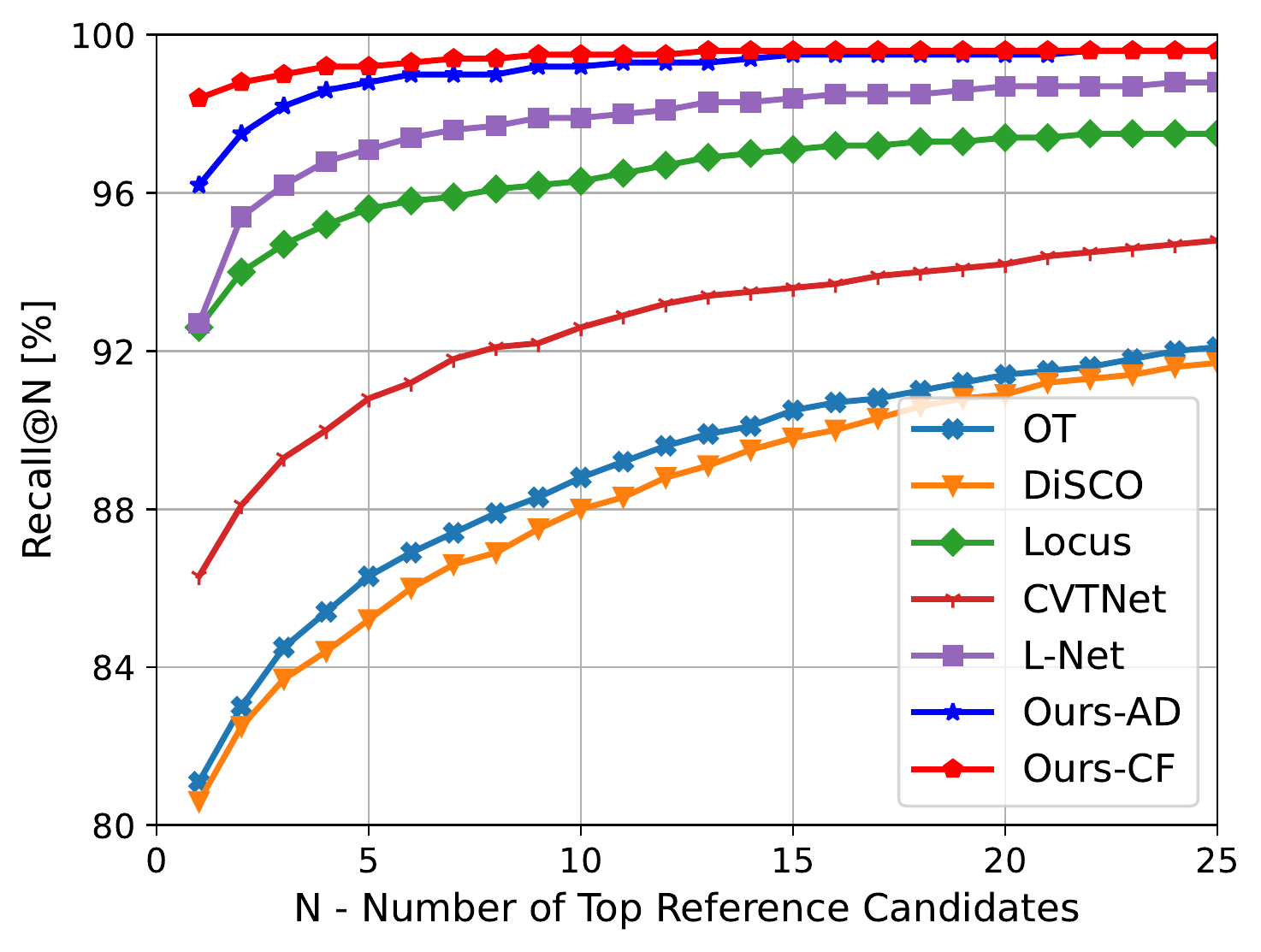}
        }
        \vspace{-3mm}
    \caption{Recall@N on KITTI and KITTI-360 datasets. }\vspace{-3mm}
    \label{pic:recall_curvs}
 \end{figure*}

\begin{table*}[h]\footnotesize
    \caption{\centering Recall@1  and Recall@1\% on KITTI and KITTI-360 datasets}\vspace{-3mm}
    \label{table:recall}
    \begin{center}    
    \begin{threeparttable}
    {
\renewcommand{\tabcolsep}{4pt}
\begin{tabular}{l|ccccccc|ccc}
\hline
\multicolumn{1}{l|}{\multirow{2}{*}{Methods}} & \multicolumn{7}{c|}{KITTI}                                                                 & \multicolumn{3}{c}{KITTI-360}     \\
\cline{2-11}
\multicolumn{1}{c|}{}                         & 00          & 02          & 05          & 06           & 07          & 08          & Mean       & 0002         & 0009         & Mean      \\
\hline
OT\cite{ma2022ral}                                          & 89.3/95.0  & 82.3/93.9  & 90.7/94.5  & 99.3/\textbf{100.0}  & 86.3/96.3  & 71.8/97.7  & 86.6/96.2  & 78.0/97.0 & 81.1/95.7 & 79.6/96.4 \\
DiSCO\cite{xu2021ral}                                     & 90.3/97.7  & 83.7/91.9  & 88.1/96.2  & 99.3/\textbf{100.0}  & 91.3/98.8  & 83.8/93.2  & 89.4/96.3  & 72.1/98.1 & 80.6/96.6 & 76.4/97.4 \\
Locus\cite{vid2021icra}                                   & 93.2/98.9  & 72.4/85.2  & 91.7/98.6  & 97.8/\textbf{100.0}  & 92.5/95.0  & 92.9/98.4  & 90.1/96.0  & 85.1/97.3 & 92.6/98.9 & 88.8/98.1 \\
CVTNet\cite{ma2023tii}                                   & 93.6/99.1  & 86.6/97.7  & 93.8/97.8  & \textbf{100.0}/\textbf{100.0} & \textbf{98.8}/\textbf{100.0} & 77.9/97.9  & 91.8/98.8  & 82.4/97.8 & 86.3/97.1 & 84.4/97.4 \\
L-Net\cite{vid2022icra}                                      & 92.5/99.1  & 91.6/96.5  & 92.9/98.1  & \textbf{100.0}/\textbf{100.0} & 91.9/98.1  & 87.5/98.9  & 92.7/98.4  & 88.5/99.4 & 92.7/99.4 & 90.6/99.4 \\
RINet\cite{li2022ral}                                      & \textbf{98.9}/\textbf{100.0} & 80.0/99.7  & 97.6/99.6  & \textbf{100.0}/\textbf{100.0} & 90.1/\textbf{100.0} & 97.4/\textbf{100.0} & 94.0/99.9  & -         & -         & -         \\
BEVNet\cite{li2023icra}                                     & 96.8/\textbf{100.0} & 98.3/\textbf{100.0} & \textbf{97.9}/\textbf{100.0} & 98.9/99.6   & 96.9/\textbf{100.0} & 97.9/\textbf{100.0} & 97.8/99.9  & -         & -         & -         \\
Ours-AD                                      & 95.9/\textbf{100.0} & 94.8/\textbf{100.0} & 95.2/99.7  & 99.6/\textbf{100.0}  & \textbf{98.8}/\textbf{100.0} & 92.3/\textbf{100.0} & 96.1/\textbf{100.0} & 94.8/\textbf{99.7} & 96.2/\textbf{99.8} & 95.5/\textbf{99.8} \\
Ours-CF                                      & 96.7/\textbf{100.0} & \textbf{98.5}/\textbf{100.0} & 97.8/99.7  & 98.9/\textbf{100.0}  & \textbf{98.8}/\textbf{100.0} & \textbf{98.2}/\textbf{100.0} & \textbf{98.1}/\textbf{100.0} & \textbf{96.6}/99.4 & \textbf{98.4}/99.6 & \textbf{97.5}/99.5 \\
\hline

\end{tabular}
    }\vspace{-6mm}
\end{threeparttable}
\end{center}
\end{table*}

\subsection{Datasets}
\textbf{KITTI.} The KITTI dataset \cite{geiger2013ijrr} contains point clouds from various urban environments, collected using a Velodyne HDL-64E 3D LiDAR sensor. The dataset includes 22 sequences, but only the first 11 sequences have ground truth poses, so we use the ground truth provided by SemanticKITTI\cite{behley2019iccv}. We train our network on the last 11 sequences and evaluate the performance of loop closing on sequences 00, 02, 05, 06, 07, and 08.

\textbf{KITTI-360.} The KITTI-360 dataset \cite{liao2022tpami} is a suburban driving dataset that consists of 9 sequences, with 6 sequences containing loops. Compared to the KITTI dataset, KITTI-360 contains more scans and more loops and reverse loops. Following \cite{cattaneo2022tro}, we evaluate our approach on sequence 0002 and sequence 0009, which contain the highest number of loop closures. The dataset contains 3D data from a Velodyne HDL-64E and a SICK LMS 200 LiDAR sensor. In our experiments, we use the 3D raw scans captured by the Velodyne scanner.

\subsection{Implementation Details} 
In the preprocessing stage, we crop each point cloud with a [-50m, 50m] cubic window and [-4m, 3m] in the z axis, then voxelize the input point cloud into a BEV representation with the shape of $256\times256\times32$. The encoder network subsequently extracts a BEV feature volume with the shape of $32\times32\times512$ using the SpConv\cite{yan2018sensors} for improved speed. These features are saved in the database $\text{DB}_f$.
 
For global descriptor generation, we use a NetVLAD layer with an intermediate feature dimension of 512, a maximum number of pooled samples of 1024, and 32 clusters. The final attention-guided global descriptor vector $v$ has a length of 1024. For triplet loss, we set the margin $m = 0.3$, $\sigma_{pos} = 10$, $\sigma_{neg} = 50$, and use 2 positive and 10 negative samples for training. The overlap estimator predicts the pairwise overlap region using a $1\times32\times32$ tensor.
 
We only train the attention-guided descriptor generation network while using the pre-trained model from \cite{li2023icra} to extract BEV features and perform overlap estimation, which is also trained on the last 11 sequences of the KITTI dataset. For candidate selection in the coarse stage, we use $K=25$ as the number of candidates. We denote our method with only global descriptor matching as ours-AD and the complete coarse-to-fine method as ours-CF.

\subsection{Evaluation of Loop Closure Detection} \label{sec_loop}
We compare our approach with state-of-the-art methods, including description-based methods: OverlapTransformer\cite{ma2022ral} (denoted as OT), DiSCO\cite{xu2021ral}, Locus\cite{vid2021icra}, CVTNet\cite{ma2023tii}, LoGG3D-Net\cite{vid2022icra} (denoted as L-Net), and pairwise similarity-based methods: RINet\cite{li2022ral} and BEVNet\cite{li2023icra}. For fairness, all these methods trained on other sequences are retrained on sequences 11-21 of the KITTI dataset, except Locus, which does not require training. During the evaluation, we exclude the previous 50 scans of the query scan as they are too close. 
 
We use Recall@1 and Recall@1\% as the evaluation metrics. The inference is considered correct if the distance between two scans is less than 10m. As shown in Tab.~\ref{table:recall} and Fig.~\ref{pic:recall_curvs}, ours-CF achieves the highest AR@1 on the KITTI dataset. Ours-CF is based on BEVNet, but we improve place recognition ability while significantly reducing the time-consuming pairwise overlap estimation process. In general, pairwise similarity-based methods perform better than description-based methods, where CVTNet achieves the best performance except for ours-AD, as it uses a cross-view transformer network to fuse the features extracted from range image views and bird's eye views images, but the result in sequence 08 shows it does not detect reverse loop closures well, as sequence 08 only has reverse loops. In terms of AR@1\%, RINet, BEVNet, Ours-AD, and Ours-CF achieve nearly 100\%. The AR@1\% of ours-AD indicates that our candidate selection process provides valid and accurate candidates, which is the foundation of our approach.

\begin{figure}[tb]
    \centering
        \includegraphics[width=\columnwidth]{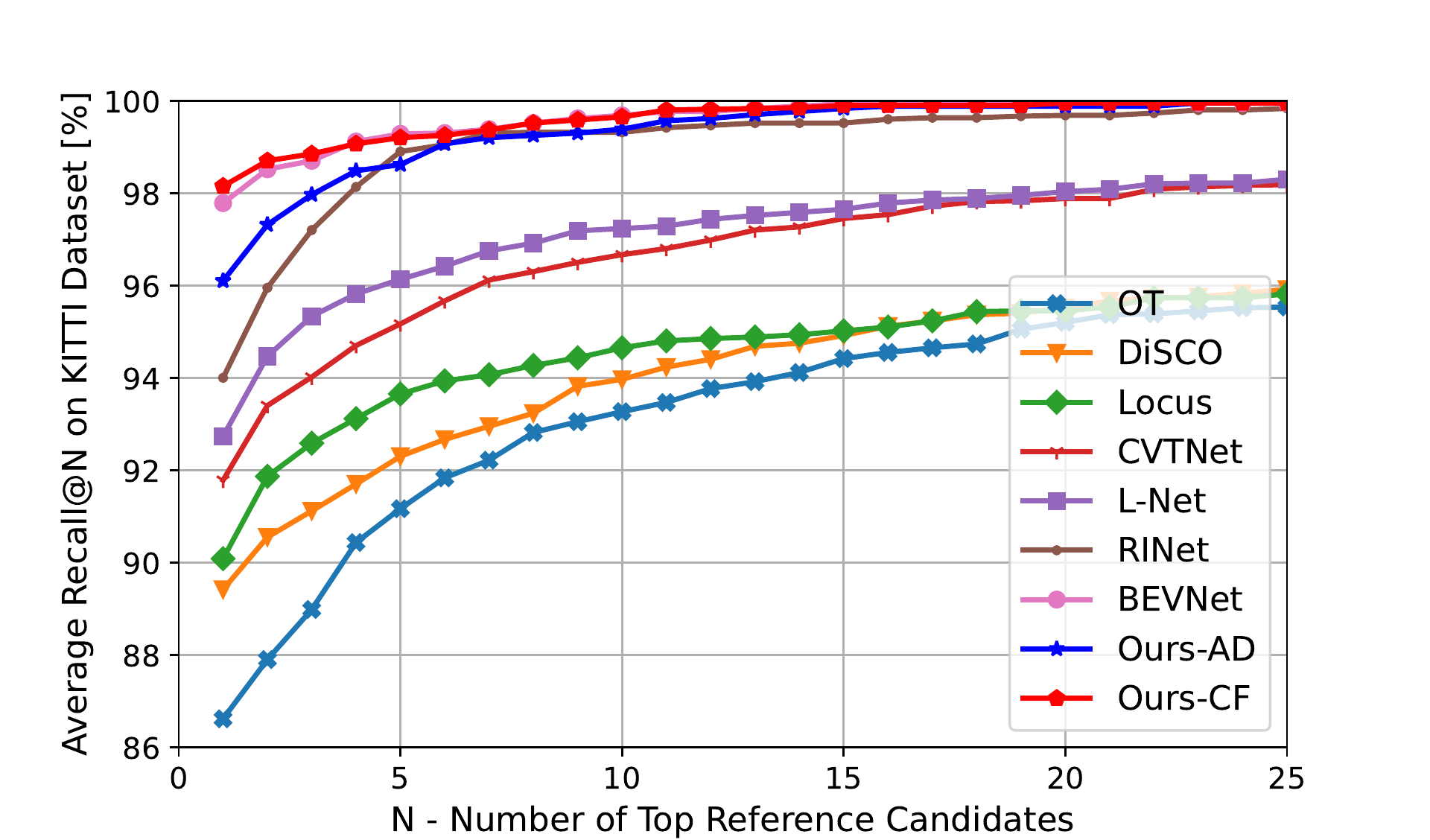}
    \caption{AR@N of ours-CF on KITTI dataset.}
    \label{pic:average_recall}
\end{figure}

Furthermore, we evaluate the performance of the proposed method using AR@N on the KITTI dataset. As shown in Fig.~\ref{pic:average_recall}, our method outperforms other methods. The experimental results demonstrate our methods' strong robustness and good generalization ability for place recognition under challenging conditions.

To assess the generalization ability of each method, we conduct direct evaluations on the KITTI-360 dataset without additional training. Since semantic annotations are provided on accumulated point clouds instead of 3D raw scans, we do not evaluate RINet on the KITTI-360 dataset. Furthermore, we do not evaluate BEVNet as it requires a significant amount of time, as discussed in detail in Sec.~\ref{sec_run_time}. It can be observed that ours-AD and ours-CF demonstrate consistent recognition ability across different datasets.

\subsection{Ablation Study} \label{sec_as}
In this section, we present our ablation studies on the proposed approach. First, we investigate the impact of the global descriptor matching in the coarse stage. We choose the number of place candidates as $N=1, 10, 15, 20, 25, 1\% $ to test Recall@N for place recognition using only global descriptor matching, which we refer to as ours-AD. As shown in Tab.~\ref{table:asa}, both AR@25 and AR@1\% are nearly 100.0\%. This result suggests that 25 candidates are sufficient to achieve outstanding performance. 

\begin{table}[tb]\footnotesize
\caption{\centering Study on candidate selection ability}\vspace{-3mm}
\label{table:asa}
\begin{center}
\begin{threeparttable}
    {
\renewcommand{\tabcolsep}{5pt}
\begin{tabular}{l c c c c c c c}
\hline
&00&02&05&06&07&08&Mean\\
\hline
Recall@1&95.9&94.8&95.2&99.6&98.8&92.3&96.1 \\
Recall@10&99.0&99.7&98.5&100.0&100.0&99.1&99.4 \\
Recall@15&99.9&100.0&99.3&100.0&100.0&99.8&99.8 \\
Recall@20&100.0&100.0&99.3&100.0&100.0&100.0&99.9 \\
Recall@25&100.0&100.0&99.7&100.0&100.0&100.0&100.0 \\
Recall@1\%&100.0&100.0&99.7&100.0&100.0&100.0&100.0 \\
\hline
\end{tabular}
}
\end{threeparttable}
\end{center}
\end{table}

Next, we further evaluate the effect of candidate number \textit{K} in the fine stage where the overlap estimation is applied. As shown in Tab.~\ref{table:asb}, Recall@1 increases with the number of candidates, as expected. Ours-CF-25 has a similar performance to ours-CF-1\%. Combining these results with those in Tab.~\ref{table:asa}, we choose $K=25$ as the candidate number for a coarse search. Besides, a fixed candidate number can help stabilize running time.

\begin{table}[tb]\footnotesize
    \caption{\centering Study on candidate number of ours-CF}\vspace{-3mm}
    \label{table:asb}
    \begin{center}
    \begin{threeparttable}
        {
    \begin{tabular}{l c c c c c c c}
    \hline
    Methods & 00 & 02 & 05 & 06 & 07 &08&Mean\\ 
    \hline
    Ours-CF-1&95.9&94.8&95.2&99.6&98.8&92.3&96.1\\
    Ours-CF-10&96.5&98.3&96.9&98.9&98.8&97.5&97.8\\
    Ours-CF-15&97.0&98.5&97.1&98.9&98.8&97.5&98.0\\
    Ours-CF-20&96.8&98.5&97.1&98.9&98.8&97.9&98.0\\
    Ours-CF-25&96.7&98.5&97.8&98.9&98.8&98.2&98.1\\
    Ours-CF-1\%&96.8&98.3&97.9&98.9&98.8&98.2&98.1\\
    \hline
    \end{tabular}
    }
    \end{threeparttable}
    \begin{tablenotes} 
        \footnotesize
        \item Ours-CF-\textit{K}\ means overlap estimation on Top-\textit{K} candidates proposed by descriptor matching. The result is Recall@1.
     \end{tablenotes}\vspace{-3mm}
\end{center}
\end{table}

We also test the impact of the self-attention module, as shown in Tab.~\ref{table:asc}, in which ours-GD uses two $3\times3$ convolutional layers and a sigmoid layer with ReLU activation instead of the self-attention module. The results clearly demonstrate that the self-attention module significantly enhances the ability to detect reverse loops.

\begin{table}[tb]\footnotesize
    \caption{\centering Study on self-attention module (Recall@1)}\vspace{-3mm}
    \label{table:asc}
    \begin{center}
    \begin{threeparttable}
\begin{tabular}{lccccccc}
\hline
Methods & 00 & 02 & 05 & 06 & 07 &08&Mean\\ 
\hline
Ours-GD & 95.9 & 91.0 & 94.3 & 99.6 & 97.5 & 84.7 & 93.8  \\
Ours-AD & 95.9 & 94.8 & 95.2 & 99.6 & 98.8 & 92.3 & 96.1 \\
\hline
\end{tabular}
    \end{threeparttable}
\end{center}
\end{table}

\subsection{Runtime}\label{sec_run_time}
In this section, we conduct experiments to evaluate the efficiency of our method. The experiments are performed on a system equipped with an Intel Core i7-7700 CPU and an Nvidia GeForce GTX 1080 Ti GPU, using sequence 00, which contains 4541 scans. First, we evaluate the runtime efficiency of attention-guided descriptor generation and compare it to other methods. The preprocessing steps include range image generation (OverlapTransformer), BEV representation construction (DiSCO and ours-AD), segments extraction (Locus), range and BEV image generation (CVTNet), and voxelization (L-Net). The batch size is set to 1 for all methods. As shown in Tab.~\ref{table:runtimeb}, the global descriptor generation in our approach is quite efficient compared to the other methods, which is the foundation of our proposed coarse-to-fine approach.

\begin{table}[t]\footnotesize
    \caption{\centering Runtime of descriptor generation (ms)}\vspace{-3mm}
    \label{table:runtimeb}
    \begin{center}
    \begin{threeparttable}
        {
    \begin{tabular}{c c c c}
    \hline
    Methods&Preprocessing &Description&Total\\
    \hline
    OT&26.92&2.22&29.14\\
    DiSCO&11.67&3.75&15.42\\
    Locus&1166.49&613.96&1780.45\\
    CVTNet&191.96&13.13&205.09 \\
    L-Net&31.64&63.88&95.52 \\
    Ours-AD&34.86&43.36&78.22\\
    \hline
    \end{tabular}
    }
    \end{threeparttable}
    \end{center}
    \end{table}

Tab.~\ref{table:runtime} shows the time consumption of each module in our proposed approach. The candidate selection only takes 0.04ms, while the pairwise overlap estimation takes 8.78ms. Thanks to fast candidate selection, ours-CF performs overlap estimation a limited number of times, whereas BEVNet uses pairwise overlap estimation to search exhaustively. In Sec.~\ref{sec_loop}, when evaluating BEVNet on sequence 0002 and sequence 0009, we encounter unacceptable long loop detection times. In sequence 0002, there are 14,122 valid scans and 4,134 loops, which would require 31,408,854 pairwise overlap estimations if exhaustively searching without candidate selection. The time cost of ours-CF is $14122*(34.86+41.74+1.62) + 4134 * (0.04 + 25 * 8.78) ms= 0.56h$. In contrast, the time cost of BEVNet is $14122*(34.86+41.74+1.62) + 8.78*31408854 ms= 76.91h$. It is evident that the time required for BEVNet to detect loop closures is significantly longer than the time for LiDAR data acquisition. In ours-CF, the speed can be further improved by adjusting the value of $K$ according to the actual needs. When $K=1$, it degenerates into ours-AD, which is still capable of detecting loops effectively. This experiment demonstrates the effectiveness and efficiency of combining descriptor matching and pairwise overlap estimation on shared BEV features.

\begin{table}[t]\footnotesize
    \caption{Runtime of different modules (ms)}\vspace{-3mm}
    \label{table:runtime}
    \begin{center}
    \begin{threeparttable}
        {
    \begin{tabular}{c c}
    \hline
    Module&Runtime\\
    \hline
    Voxelization&34.86\\
    BEV Feature Extraction&41.74\\
    Attention-guided Descriptor Generation&1.62\\
    Affinity-based Candidate Selection&0.04\\
    Pairwise Overlap Estimation&8.78\\
    \hline
    \end{tabular}\vspace{-3mm}
    }
    \end{threeparttable}
    \end{center}
    \end{table}

\section{CONCLUSION}
In this paper, we propose a novel coarse-to-fine approach for LiDAR-based place recognition. This approach combines global descriptor matching and overlap estimation based on shared BEV features to achieve both speed and accuracy in loop closure detection. While our approach has some limitations, such as the bulky and memory-intensive BEV features, future work will address these challenges and focus on improving efficiency and seeking better results.

\bibliographystyle{ieeetr}
\bibliography{main}

\begin{thebibliography}{10}

\bibitem{uy2018cvpr}
M.~A. Uy and G.~H. Lee, ``Pointnetvlad: Deep point cloud based retrieval for
  large-scale place recognition,'' in {\em Proceedings of the IEEE conference
  on computer vision and pattern recognition}, pp.~4470--4479, 2018.

\bibitem{vid2022icra}
K.~Vidanapathirana, M.~Ramezani, P.~Moghadam, S.~Sridharan, and C.~Fookes,
  ``Logg3d-net: Locally guided global descriptor learning for 3d place
  recognition,'' in {\em 2022 International Conference on Robotics and
  Automation (ICRA)}, pp.~2215--2221, 2022.

\bibitem{komorowski2022icpr}
J.~Komorowski, ``Improving point cloud based place recognition with
  ranking-based loss and large batch training,'' in {\em 2022 26th
  International Conference on Pattern Recognition (ICPR)}, pp.~3699--3705,
  2022.

\bibitem{komo2021wacv}
J.~Komorowski, ``Minkloc3d: Point cloud based large-scale place recognition,''
  in {\em Proceedings of the IEEE/CVF Winter Conference on Applications of
  Computer Vision}, pp.~1790--1799, 2021.

\bibitem{kong2020iros}
X.~Kong, X.~Yang, G.~Zhai, X.~Zhao, X.~Zeng, M.~Wang, Y.~Liu, W.~Li, and
  F.~Wen, ``Semantic graph based place recognition for 3d point clouds,'' in
  {\em 2020 IEEE/RSJ International Conference on Intelligent Robots and Systems
  (IROS)}, pp.~8216--8223, IEEE, 2020.

\bibitem{vid2021icra}
K.~Vidanapathirana, P.~Moghadam, B.~Harwood, M.~Zhao, S.~Sridharan, and
  C.~Fookes, ``Locus: Lidar-based place recognition using spatiotemporal
  higher-order pooling,'' in {\em IEEE International Conference on Robotics and
  Automation (ICRA)}, 2021.

\bibitem{ma2022ral}
J.~Ma, J.~Zhang, J.~Xu, R.~Ai, W.~Gu, and X.~Chen, ``Overlaptransformer: An
  efficient and yaw-angle-invariant transformer network for lidar-based place
  recognition,'' {\em IEEE Robotics and Automation Letters}, vol.~7, no.~3,
  pp.~6958--6965, 2022.

\bibitem{ma2022tie}
J.~Ma, X.~Chen, J.~Xu, and G.~Xiong, ``Seqot: A spatial-temporal transformer
  network for place recognition using sequential lidar data,'' {\em IEEE
  Transactions on Industrial Electronics}, pp.~1--10, 2022.

\bibitem{yin2020iros}
P.~Yin, F.~Wang, A.~Egorov, J.~Hou, J.~Zhang, and H.~Choset, ``Seqspherevlad:
  Sequence matching enhanced orientation-invariant place recognition,'' in {\em
  2020 IEEE/RSJ International Conference on Intelligent Robots and Systems
  (IROS)}, pp.~5024--5029, 2020.

\bibitem{zhao2023ral}
S.~Zhao, P.~Yin, G.~Yi, and S.~Scherer, ``Spherevlad++: Attention-based and
  signal-enhanced viewpoint invariant descriptor,'' {\em IEEE Robotics and
  Automation Letters}, vol.~8, no.~1, pp.~256--263, 2023.

\bibitem{wang2020iros}
Y.~Wang, Z.~Sun, C.-Z. Xu, S.~E. Sarma, J.~Yang, and H.~Kong, ``Lidar iris for
  loop-closure detection,'' in {\em 2020 IEEE/RSJ International Conference on
  Intelligent Robots and Systems (IROS)}, pp.~5769--5775, IEEE, 2020.

\bibitem{kim2018iros}
G.~Kim and A.~Kim, ``Scan context: Egocentric spatial descriptor for place
  recognition within 3d point cloud map,'' in {\em 2018 IEEE/RSJ International
  Conference on Intelligent Robots and Systems (IROS)}, pp.~4802--4809, IEEE,
  2018.

\bibitem{kim2021tro}
G.~Kim, S.~Choi, and A.~Kim, ``Scan context++: Structural place recognition
  robust to rotation and lateral variations in urban environments,'' {\em IEEE
  Transactions on Robotics}, vol.~38, no.~3, pp.~1856--1874, 2021.

\bibitem{wang2020icra}
H.~Wang, C.~Wang, and L.~Xie, ``Intensity scan context: Coding intensity and
  geometry relations for loop closure detection,'' in {\em 2020 IEEE
  International Conference on Robotics and Automation (ICRA)}, pp.~2095--2101,
  2020.

\bibitem{li2021iros}
L.~Li, X.~Kong, X.~Zhao, T.~Huang, W.~Li, F.~Wen, H.~Zhang, and Y.~Liu, ``Ssc:
  Semantic scan context for large-scale place recognition,'' in {\em 2021
  IEEE/RSJ International Conference on Intelligent Robots and Systems (IROS)},
  pp.~2092--2099, IEEE, 2021.

\bibitem{xu2021ral}
X.~Xu, H.~Yin, Z.~Chen, Y.~Li, Y.~Wang, and R.~Xiong, ``Disco: Differentiable
  scan context with orientation,'' {\em IEEE Robotics and Automation Letters},
  vol.~6, no.~2, pp.~2791--2798, 2021.

\bibitem{luo2021ral}
L.~Luo, S.-Y. Cao, B.~Han, H.-L. Shen, and J.~Li, ``Bvmatch: Lidar-based place
  recognition using bird's-eye view images,'' {\em IEEE Robotics and Automation
  Letters}, vol.~6, no.~3, pp.~6076--6083, 2021.

\bibitem{jiang2023icra}
B.~Jiang and S.~Shen, ``Contour context: Abstract structural distribution for
  3d lidar loop detection and metric pose estimation,'' in {\em 2023 IEEE
  International Conference on Robotics and Automation (ICRA)}, (London, United
  Kingdom), p.~8386–8392, 2023.

\bibitem{luo2023}
L.~Lun, Z.~Shuhang, L.~Yixuan, F.~Yongzhi, Y.~Beinan, C.~Siyuan, and
  S.~Hui-Liang, ``{BEVPlace}: {Learning LiDAR-based} place recognition using
  bird's eye view images,'' {\em arXiv preprint arXiv:2302.14325}, 2023.

\bibitem{chen2020rss}
X.~Chen, T.~L\"abe, A.~Milioto, T.~R\"ohling, O.~Vysotska, A.~Haag, J.~Behley,
  and C.~Stachniss, ``{OverlapNet: Loop Closing for LiDAR-based SLAM},'' in
  {\em Proceedings of Robotics: Science and Systems (RSS)}, 2020.

\bibitem{li2022ral}
L.~Li, X.~Kong, X.~Zhao, T.~Huang, W.~Li, F.~Wen, H.~Zhang, and Y.~Liu,
  ``Rinet: Efficient 3d lidar-based place recognition using rotation invariant
  neural network,'' {\em IEEE Robotics and Automation Letters}, vol.~7, no.~2,
  pp.~4321--4328, 2022.

\bibitem{li2023icra}
L.~Li, W.~Ding, Y.~Wen, Y.~Liang, Y.~Liu, and G.~Wan, ``A unified bev model for
  joint learning of 3d local features and overlap estimation,'' {\em arXiv
  preprint arXiv:2302.14511}, 2023.

\bibitem{geiger2013ijrr}
A.~Geiger, P.~Lenz, C.~Stiller, and R.~Urtasun, ``Vision meets robotics: The
  kitti dataset,'' {\em The International Journal of Robotics Research},
  vol.~32, no.~11, pp.~1231--1237, 2013.

\bibitem{liao2022tpami}
Y.~Liao, J.~Xie, and A.~Geiger, ``{KITTI}-360: A novel dataset and benchmarks
  for urban scene understanding in 2d and 3d,'' {\em Pattern Analysis and
  Machine Intelligence (PAMI)}, 2022.

\bibitem{chen2017icra}
Z.~Chen, A.~Jacobson, N.~S{\"u}nderhauf, B.~Upcroft, L.~Liu, C.~Shen, I.~Reid,
  and M.~Milford, ``Deep learning features at scale for visual place
  recognition,'' in {\em 2017 IEEE international conference on robotics and
  automation (ICRA)}, pp.~3223--3230, IEEE, 2017.

\bibitem{gehrig2017icra}
M.~Gehrig, E.~Stumm, T.~Hinzmann, and R.~Siegwart, ``Visual place recognition
  with probabilistic voting,'' in {\em 2017 IEEE International Conference on
  Robotics and Automation (ICRA)}, pp.~3192--3199, IEEE, 2017.

\bibitem{schubert2020icra}
S.~Schubert, P.~Neubert, and P.~Protzel, ``Unsupervised learning methods for
  visual place recognition in discretely and continuously changing
  environments,'' in {\em 2020 IEEE International Conference on Robotics and
  Automation (ICRA)}, pp.~4372--4378, 2020.

\bibitem{camara2020icra}
L.~G. Camara, C.~Gäbert, and L.~Přeučil, ``Highly robust visual place
  recognition through spatial matching of cnn features,'' in {\em 2020 IEEE
  International Conference on Robotics and Automation (ICRA)}, pp.~3748--3755,
  May 2020.

\bibitem{zhang2021icra}
K.~Zhang, Z.~Li, and J.~Ma, ``Appearance-based loop closure detection via
  bidirectional manifold representation consensus,'' in {\em 2021 IEEE
  International Conference on Robotics and Automation (ICRA)}, pp.~6811--6817,
  IEEE, 2021.

\bibitem{he2016iros}
L.~He, X.~Wang, and H.~Zhang, ``M2dp: A novel 3d point cloud descriptor and its
  application in loop closure detection,'' in {\em 2016 IEEE/RSJ International
  Conference on Intelligent Robots and Systems (IROS)}, pp.~231--237, 2016.

\bibitem{qi2017cvpr}
C.~R. Qi, H.~Su, K.~Mo, and L.~J. Guibas, ``Pointnet: Deep learning on point
  sets for 3d classification and segmentation,'' in {\em Proceedings of the
  IEEE conference on computer vision and pattern recognition}, pp.~652--660,
  2017.

\bibitem{arandjelovic2016cvpr}
R.~Arandjelovic, P.~Gronat, A.~Torii, T.~Pajdla, and J.~Sivic, ``Netvlad: Cnn
  architecture for weakly supervised place recognition,'' in {\em Proceedings
  of the IEEE conference on computer vision and pattern recognition},
  pp.~5297--5307, 2016.

\bibitem{ma2023tii}
J.~Ma, G.~Xiong, J.~Xu, and X.~Chen, ``Cvtnet: A cross-view transformer network
  for place recognition using lidar data,'' {\em arXiv preprint
  arXiv:2302.01665}, 2023.

\bibitem{cop2018icra}
K.~P. Cop, P.~V. Borges, and R.~Dub{\'e}, ``Delight: An efficient descriptor
  for global localisation using lidar intensities,'' in {\em 2018 IEEE
  International Conference on Robotics and Automation (ICRA)}, pp.~3653--3660,
  IEEE, 2018.

\bibitem{zhang2019icml}
H.~Zhang, I.~Goodfellow, D.~Metaxas, and A.~Odena, ``Self-attention generative
  adversarial networks,'' in {\em International conference on machine
  learning}, pp.~7354--7363, PMLR, 2019.

\bibitem{behley2019iccv}
J.~Behley, M.~Garbade, A.~Milioto, J.~Quenzel, S.~Behnke, C.~Stachniss, and
  J.~Gall, ``Semantickitti: A dataset for semantic scene understanding of lidar
  sequences,'' in {\em Proceedings of the IEEE/CVF international conference on
  computer vision}, pp.~9297--9307, 2019.

\bibitem{cattaneo2022tro}
D.~Cattaneo, M.~Vaghi, and A.~Valada, ``Lcdnet: Deep loop closure detection and
  point cloud registration for lidar slam,'' {\em IEEE Transactions on
  Robotics}, pp.~1--20, 2022.

\bibitem{yan2018sensors}
Y.~Yan, Y.~Mao, and B.~Li, ``Second: Sparsely embedded convolutional
  detection,'' {\em Sensors}, vol.~18, no.~10, p.~3337, 2018.

\end{thebibliography}

\end{document}